\documentclass[review]{elsarticle}

\journal{Computers and Electronics in Agriculture}

\usepackage{hyperref}   
\usepackage{amsfonts}   
\usepackage{amsmath}    
\usepackage{graphics}   
\usepackage{caption}    
\usepackage{subfig}     
\usepackage{pdflscape}  
\usepackage{multirow}   

\begin{document}
\begin{frontmatter}

\title{Visual Identification of Individual Holstein-Friesian Cattle via Deep Metric Learning}

\author[add1,add2]{William Andrew}
\author[add1]{Jing Gao}
\author[add2]{Siobhan Mullan}
\author[add1]{\\Neill Campbell}
\author[add2,add3]{Andrew W Dowsey\corref{cor1}}
\ead{andrew.dowsey@bristol.ac.uk}
\author[add1]{Tilo Burghardt}

\cortext[cor1]{Corresponding author}

\address[add1]{Department of Computer Science, Merchant Venturers Building, Woodland Road, Bristol, BS8 1UB}
\address[add2]{Bristol Veterinary School, Langford House, Bristol, BS40 5DU}
\address[add3]{Department of Population Health Sciences, Oakfield House, Oakfield Grove, Bristol, BS8 2BN}


\begin{abstract}
Holstein-Friesian cattle exhibit individually-characteristic black and white coat patterns visually akin to those arising from Turing's reaction-diffusion systems. 
This work takes advantage of these natural markings in order to automate visual detection and biometric identification of individual Holstein-Friesians via convolutional neural networks and deep metric learning techniques. 
Existing approaches rely on markings, tags or wearables with a variety of maintenance requirements, whereas we present a totally hands-off method for the automated detection, localisation, and identification of individual animals from overhead imaging in an open herd setting, \textit{i.e.} where new additions to the herd are identified without re-training. 
We find that deep metric learning systems show strong performance even when many cattle unseen during system training are to be identified and re-identified -- achieving $93.8\%$ accuracy when trained on just half of the population.
This work paves the way for facilitating the non-intrusive monitoring of cattle applicable to precision farming and surveillance for automated productivity, health and welfare monitoring, and to veterinary research such as behavioural analysis, disease outbreak tracing, and more.
Key parts of the source code, network weights and datasets are available publicly \cite{sourcecodedatasets}.
\end{abstract}


\end{frontmatter}



\section{Introduction}

\begin{figure*}[!b]
\centering
    \includegraphics[width=1.0\textwidth]{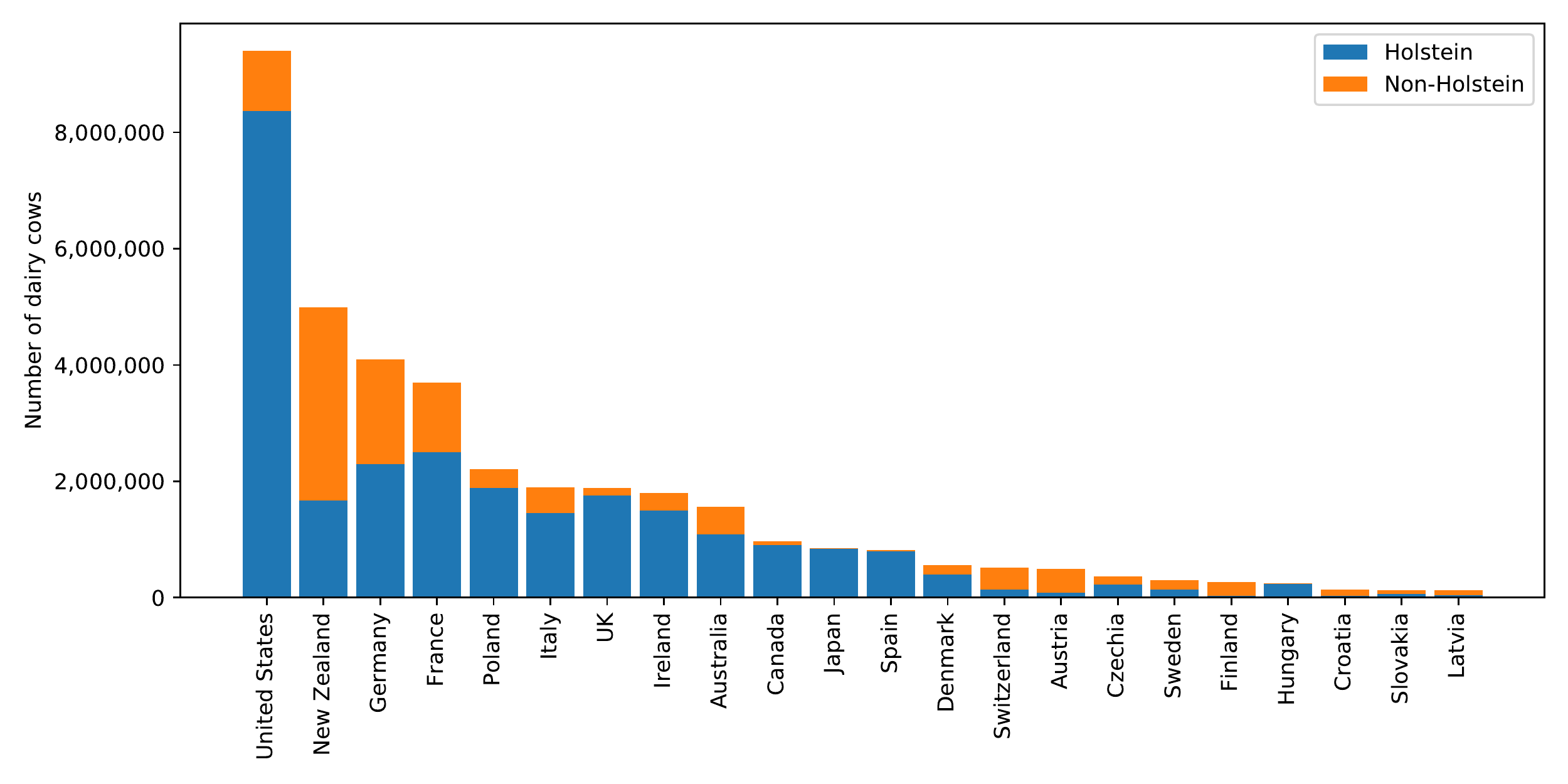}
    \caption{\textbf{Cattle breed distribution.} The numbers of Black and White Holstein and other breeds in countries reported by the World Holstein Friesian Federation \cite{whff}.}
	\label{fig:cattle-distributions}
\end{figure*}

\textbf{\textit{Motivation}}. Driven by their high milk yield \cite{tadesse2003milk}, black and white patterned Holstein-Friesian cattle are the most common and widespread cattle breed in the world, constituting around 70 million animals in 150 countries \cite{faocattle}. 
Figure \ref{fig:cattle-distributions} illustrates that they account for over 89\% of the 10 million dairy cows in North America, 63\% of Europe's 21 million dairy cows and 42\% of the 6 million dairy cows found in Australia and New Zealand \cite{faostat, whff}.
Other animals may also have patterned coats, for example those related to Holstein Friesians such as the British Friesian, as well as some of their cross breeds, for example the Girolando that produces 80\% of the milk in Brazil, a country with the second largest population of dairy cows - approximately 16 million \cite{faostat}.
Other breeds commonly have patterned coats, for example the Normande, Shorthorn and Guernsey, although these have not had the global dominance of the Holstein Friesian.
In addition, many patterned coat animals, especially males, are reared for meat.

\begin{figure*}[!b]
    \centering
    \subfloat[Ear tag]{\includegraphics[width=0.365\textwidth]{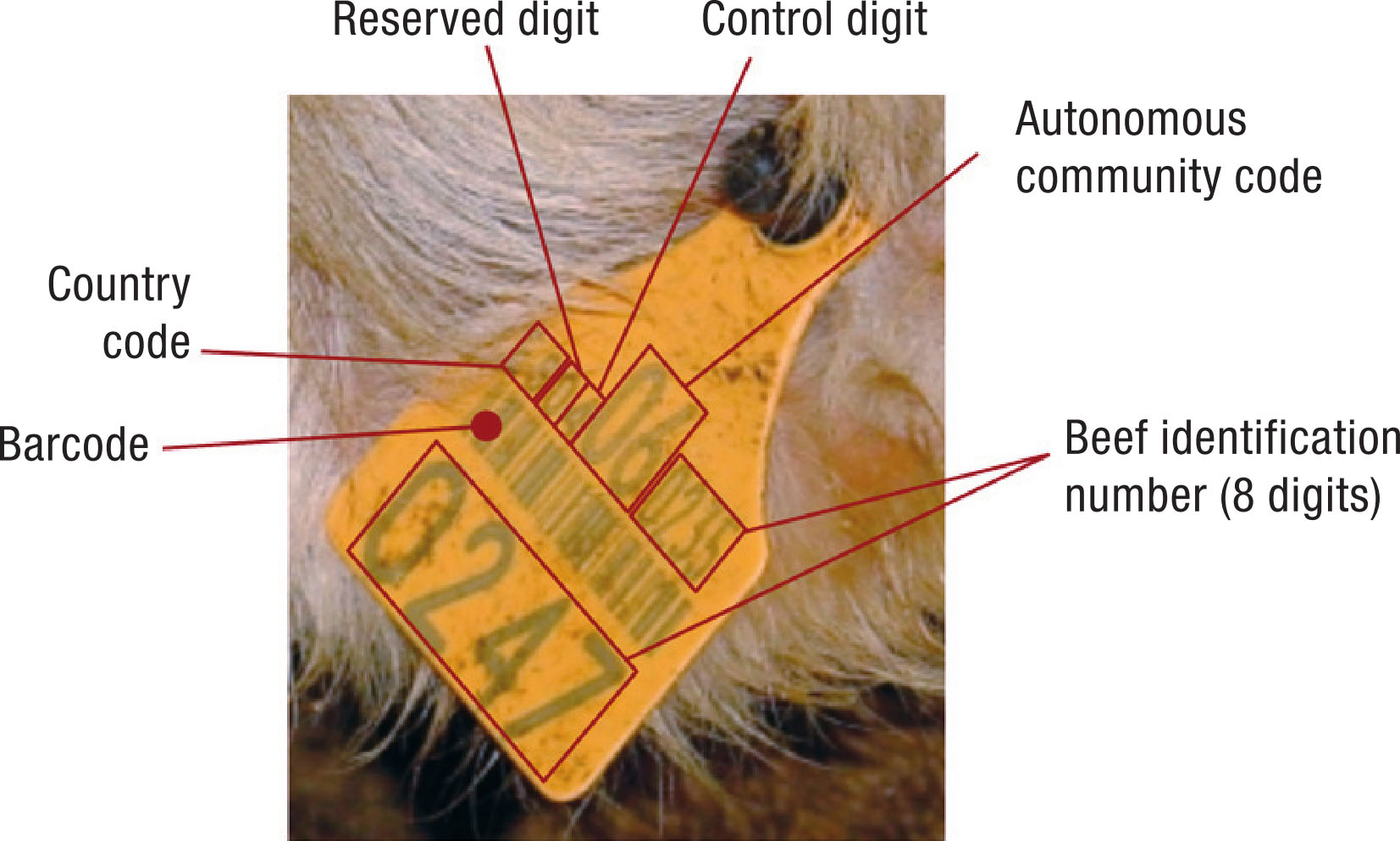}}
    \hfill
    \subfloat[Tattooing]{\includegraphics[width=0.32\textwidth]{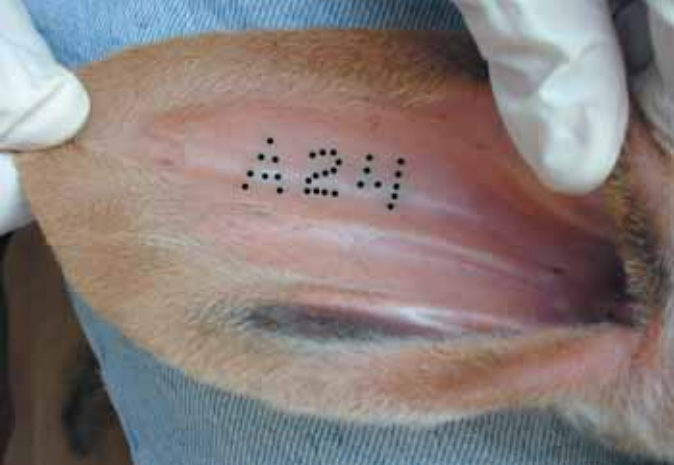}}
    \hfill
    \subfloat[Collars]{\includegraphics[width=0.295\textwidth]{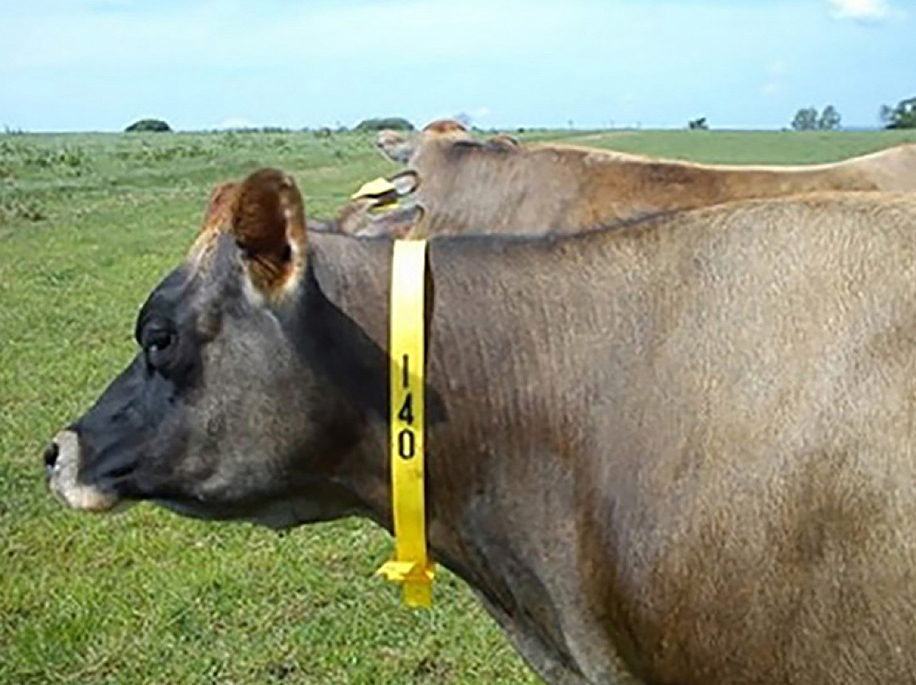}}
    \\
    \subfloat[Freeze branding]{\includegraphics[width=0.32\textwidth]{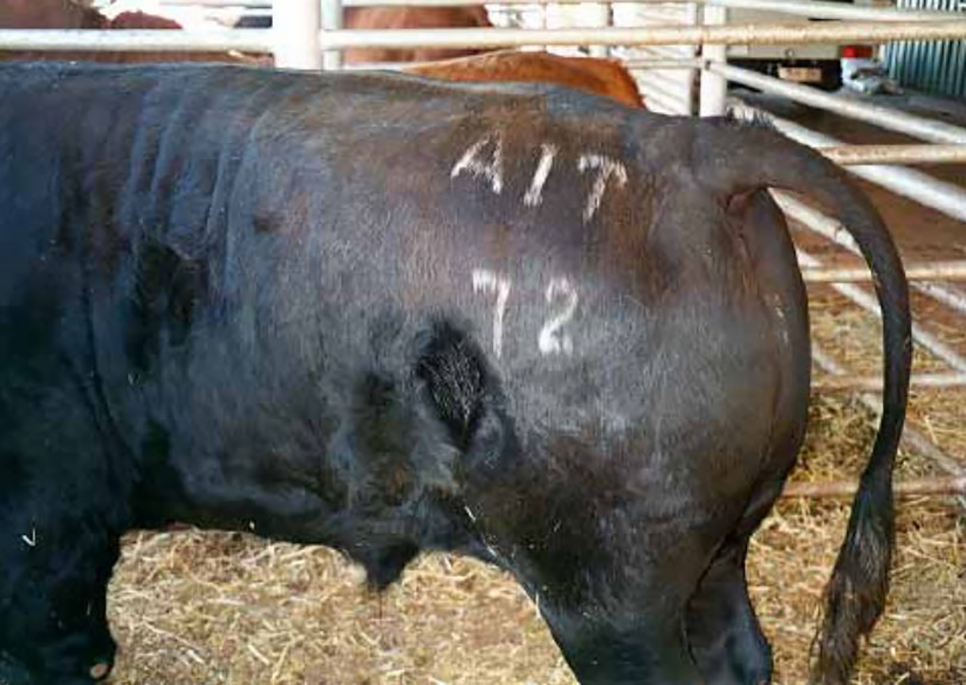}}
    \hspace{2pt}
    \subfloat[Ours]{\includegraphics[width=0.305\textwidth]{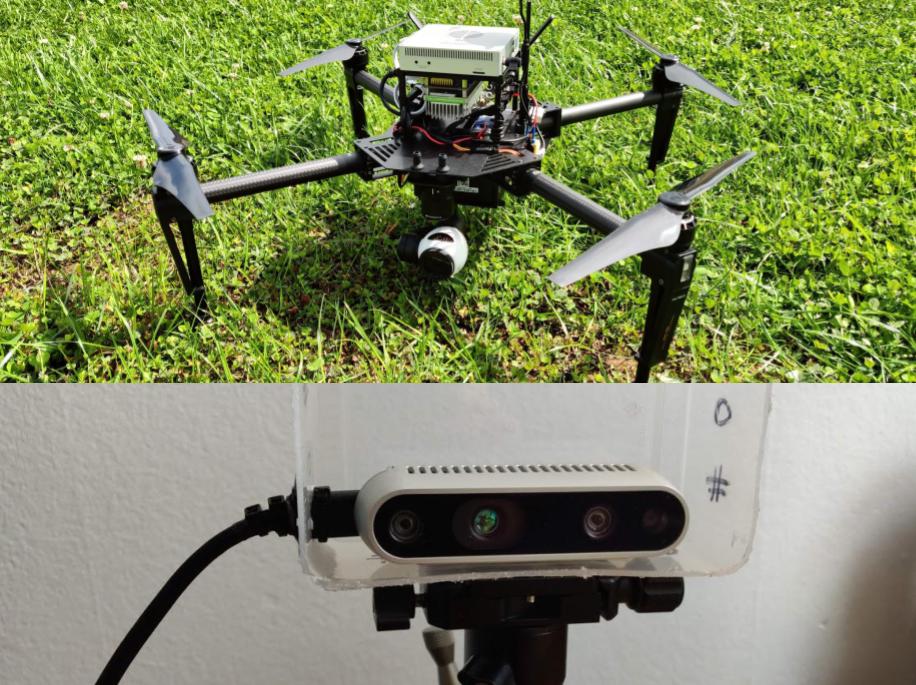}}
    \caption{\textbf{Cattle Identification Methods.} Examples of traditional methods for identifying cattle. All rely upon some physical addition, be that permanent (branding, tattooing, ear tagging) or non-permanent (collars). We instead propose to use naturally-occurring coat pattern features to achieve vision-based identification from imagery acquired via \textit{(d)} an Unmanned Aerial Vehicle (UAV) (top), or low-cost static cameras (bottom). Figure credit: \textit{(a)} Velez, J. F. et al. \cite{velez2013beef}, \textit{(b)} Pennington, J. A. \cite{pennington2007tattooing}, \textit{(c)} PyonProducts, \textit{(d)} Bertram, J. et al. \cite{bertram1996freeze}}
    \label{fig:tagging-methods}
\end{figure*}

\begin{figure}[h]
\centering
\includegraphics[width=1.0\textwidth]{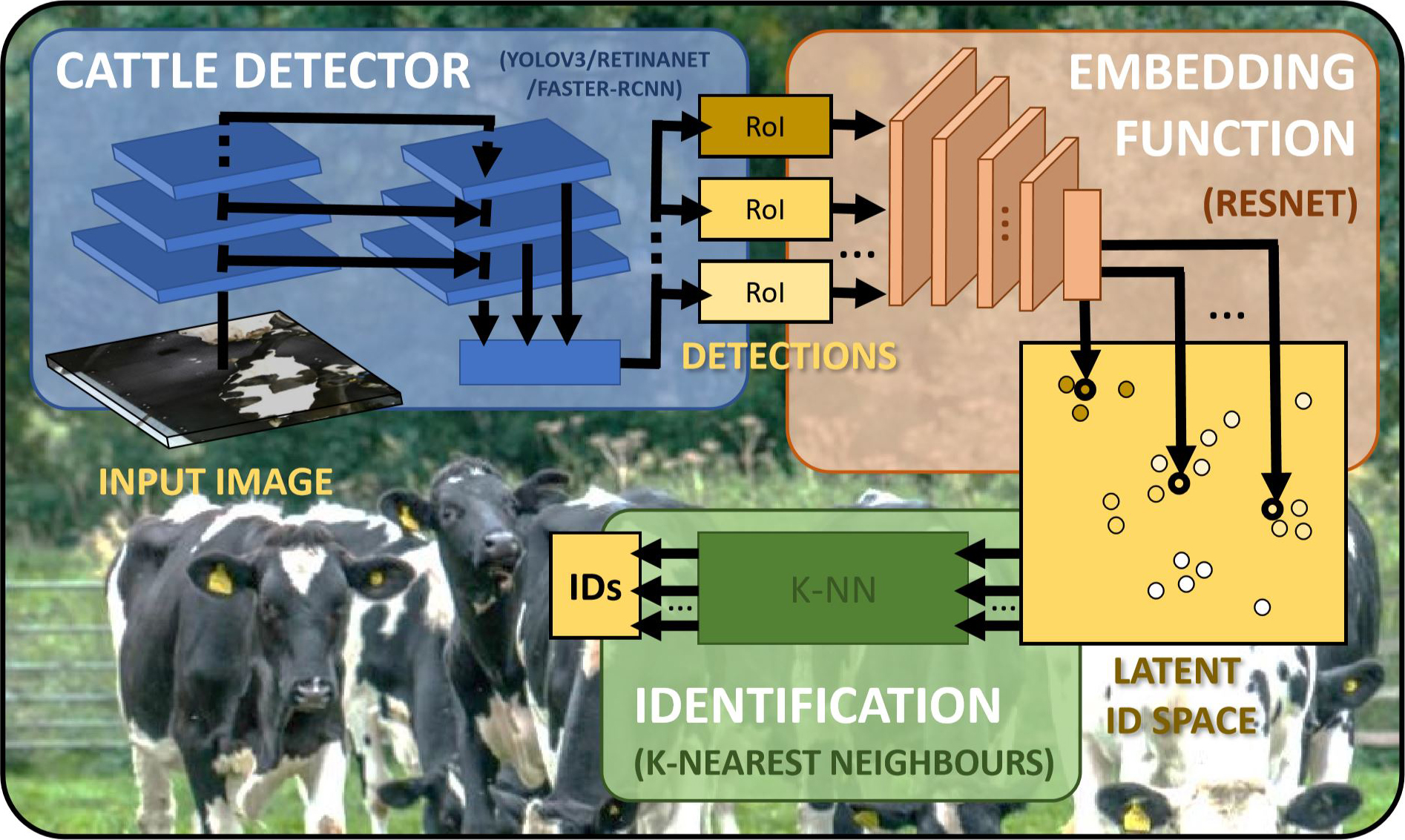}
\caption{\textbf{Identification Pipeline Overview.} Overview of the proposed pipeline for automatically detecting and identifying both known and never before seen cattle. The process begins with a breed-wide detector extracting cattle regions of interest (RoIs) agnostic to individual patterns. These are then embedded via a ResNet-driven dimensionality reduction model trained to cluster images according to individual coat patterns. RoIs projected into this latent ID space can then be classified by lightweight approaches such as k-nearest neighbours, ultimately yielding cattle identities for input images. Unknown cattle can be projected into this same space as long as the model has learnt a sufficiently discriminative reduction such that its new embeddings can be differentiated from other clusters based on distance.}
\label{fig:pipeline-overview}
\end{figure}

Legal frameworks in many countries mandate traceability of livestock throughout their lives \cite{eu82097, united2018states} in order to identify individuals for monitoring, control of disease outbreak, and more \cite{hansen2018automated, smith2005traceability, bowling2008identification, caporale2001importance}.
For cattle this is often realised in the form of a national tracking database linked to a unique ear-tag identification for each  animal \cite{houston2001computerised, buick2004animal, shanahan2009framework}, or via injectable transponders \cite{klindtworth1999electronic}, branding \cite{adcock2018branding}, and more \cite{awad2016classical} (see Fig. \ref{fig:tagging-methods}). 
Such tags, however, cannot provide the continuous localisation of individuals that would open up numerous applications in precision farming and a number of research areas including welfare assessment, behavioural and social analysis, disease development and infection transmission, amongst others~\cite{ungar2005inference, turner2000monitoring}. 
Even for conventional identification tagging has been called into question from a welfare standpoint~\cite{johnston19961418001, edwards1999welfare} regarding longevity/reliability~\cite{fosgate2006ear} and permanent damage~\cite{edwards2001comparison, wardrope1995problems}. 
Building upon previous research~\cite{martinez2013video, li2017automatic, andrew2016automatic,andrew2017visual, andrew2019aerial, andrew2019visual}, we propose to take advantage of the intrinsic, characteristic formations of the breed's coat pattern in order to perform non-intrusive visual identification (ID)~\cite{kuhl2013}, laying down the essential precursors to continuous monitoring of herds on an individual animal level via non-intrusive visual observation (see Fig.~\ref{fig:pipeline-overview}).

\textbf{\textit{Closed-set Identification.}} Our previous works showed that visual cattle detection, localisation, and re-identification via deep learning is robustly feasible in closed-set scenarios where a system is trained and tested on a fixed set of known Holstein-Friesian cattle under study~\cite{andrew2016automatic, andrew2017visual, andrew2019visual}.
However in this setup, imagery of \textit{all} animals must be taken and manually annotated/identified before system training can take place. 
Consequently, any change in the population or transfer of the system to a new herd requires labour-intensive data gathering and labelling, plus computationally demanding retraining of the system.

\textbf{\textit{Open-set Identification}.} In this paper our focus is on a more flexible scenario - the open-set recognition of individual Holstein-Friesian cattle. Instead of only being able to recognise individuals that have been seen before and trained against, the system should be able to identify and re-identify cattle that have never been seen before without further retraining. To provide a complete process, we propose a full pipeline for detection and open-set recognition from image input to IDs. (see Fig.~\ref{fig:pipeline-overview}).

The remainder of this paper and its contributions are organised as follows: Section \ref{sec:related-work} discusses relevant related works in the context of this paper.
Next, Section \ref{sec:dataset} describes the dataset used and published alongside this paper.
Section \ref{sec:detection} then outlines Holstein-Friesian breed RoI detection, the first stage of the proposed identification pipeline, followed by the second stage in Section \ref{sec:open-ID} on open-set individual recognition with extensive experiments on various relevant techniques given in Section \ref{sec:experiments}.
Finally, concluding remarks and possible avenues for future work are given in Section \ref{sec:conclusion}.


\section{Related Work}
\label{sec:related-work}

The most longstanding approaches to cattle biometrics leverage the discovery of the cattle muzzle as a dermatoglyphic trait as far back as 1922 by Petersen, WE. \cite{petersen1922identification}.
Since then, this property has been taken advantage of in the form of semi-automated approaches \cite{kumar2017automatic, kumar2017real, kimura2004structural, tharwat2014cattle} and those operating on muzzle images \cite{awad2019bag, el2015bovines, barry2007using}.
These techniques, however, rely upon the presence of heavily constrained images of the cattle muzzle that are not easily attainable.
Other works have looked towards retinal biometrics \cite{allen2008evaluation}, facial features \cite{barbedo2019study, cai2013cattle}, and body scans \cite{arslan20143d}, all requiring specialised imaging.


\subsection{Automated Cattle Biometrics}

Only a few works have utilised advancements in the field of computer vision for the automated extraction of individual identity based on full body dorsal features \cite{martinez2013video, li2017automatic}.
Our previous works have taken advantage of this property; exploiting manually-delineated features extracted on the coat \cite{andrew2016automatic} (similar to a later work by Li, W. et al. \cite{li2017automatic}), which was outperformed by a deep learning approach using convolutional neural networks extracting features from entire image sequences \cite{andrew2017visual, andrew2019aerial, andrew2019visual}, similar to \cite{hu2020cow, qiao2019individual}.
More recently, there have been works that integrate multiple views of cattle faces for identification \cite{barbedo2019study}, utilise thermal imagery for background subtraction as a pre-processing technique for a standard CNN-based classification pipeline \cite{bhole2019computer}, and detect cattle presence from UAV-acquired imagery \cite{barbedo2019study}.
In this work we continue to exploit dorsal biometric features from coat patterns exhibited by Holstein and Holstein-Friesian breeds as they provably provide sufficient distinction across populations. 
In addition, the images are easily acquired via static ceiling-mounted cameras, or outdoors using UAVs. 
Note that such birds-eye view images provide a canonical and consistent viewpoint of the object, the possibility of occlusions is widely eradicated, and imagery can be captured in a non-intrusive manner.


\subsection{Deep Object Detection}
\label{sec:review-object-detection}

Object detectors generally fall into two classes: one-stage detectors such as SSD~\cite{liu2016ssd} and  YOLO~\cite{Redmon_2016_CVPR}, which infer class probability and bounding box offsets within a single feed-forward network, and two-stage detectors such as Faster R-CNN \cite{ren2015faster} and Cascade-RCNN~\cite{cai2018cascade} that pre-process the images first to generate class-agnostic regions before classifying these and regressing associated bounding boxes. 
Recent improvements to one-stage detectors exemplified by YOLOv3~\cite{redmon2018yolov3} and RetinaNet~\cite{lin2017focal} deliver detection accuracy comparable to two-stage detectors at the general speed of a single detection stage. 
A RetinaNet architecture is used as the detection network of choice in this work, since it also addresses class imbalances; replacing the traditional cross-entropy loss with focal loss for classification.


\subsection{Open-Set Recognition}
\label{sec:review-open-set}
The problem of open-set recognition -- that is, automatically re-identifying never before seen objects -- is a well-studied area in computer vision and machine learning.
Traditional and seminal techniques typically have their foundations in probabilistic and statistical approaches~\cite{jain2014multi, scheirer2014probability, rudd2017extreme}, with alternatives including specialised support vector machines~\cite{scheirer2012toward, junior2016specialized} and others~\cite{bendale2015towards, junior2017nearest}.

However, given the performance gains on benchmark datasets achieved using deep learning and neural network techniques~\cite{sermanet2013overfeat, girshick2014rich, krizhevsky2012imagenet}, approaches to open-set recognition have followed suit.
Proposed deep models can be found to operate in an autoencoder paradigm~\cite{oza2019deep, yoshihashi2019classification}, where a network learns to transform an image input into an efficient latent representation and then reconstructs it from that representation as closely as possible.
Recent work includes the use of open-set loss function formulations instead of softmax~\cite{bendale2016towards}, the generation of counterfactual images close to the training set to strengthen object discrimination~\cite{neal2018open}, and approaches that combine these two techniques~\cite{ge2017generative, shu2017doc}.
Some additional but less relevant techniques are discussed in~\cite{geng2018recent}.

The approach taken in this work is to learn a latent representation of the training set of individual cattle in the form of an embedding that generalises visual uniqueness of the breed beyond that of the specific training herd.
The idea is that this dimensionality reduction should be discriminative to the extent that different unseen individuals projected into this space will differ significantly from the embeddings of the known training set and also each other.
This form of approach has history in the literature~\cite{meyer2019importance, lagunes2019learning, hassen2018learning}, where embeddings have been originally used for human re-identification~\cite{schroff2015facenet, hermans2017defense} as well as data aggregation and clustering~\cite{oh2016deep, opitz2018deep, oh2017deep}.
In our experiments, we will investigate the effect of various loss functions for constructing metric latent spaces~\cite{schroff2015facenet, lagunes2019learning, masullo2019goes} and quantify their suitability for the open-set recognition of Holstein-Friesian cattle.


\section{Dataset: OpenCows2020}
\label{sec:dataset}

To facilitate the experiments carried out in this paper, we introduce the OpenCows2020 dataset, which is available publicly \cite{sourcecodedatasets}.
The dataset consists of indoor and outdoor top-down imagery bringing together multiple previous works and datasets~\cite{andrew2016automatic, andrew2017visual, andrew2019aerial}.
Indoor footage was acquired with statically affixed cameras, whilst outdoor imagery was captured onboard a UAV.
The dataset is split into two components detailed below, for \textit{(a)} cattle detection and localisation, the first stage of our pipeline, and \textit{(b)} open-set identification, the second stage.


\subsection{Detection and Localisation}
\label{subsec:dataset:detection}

\begin{figure}[t]
    \centering
    \includegraphics[width=1.0\textwidth]{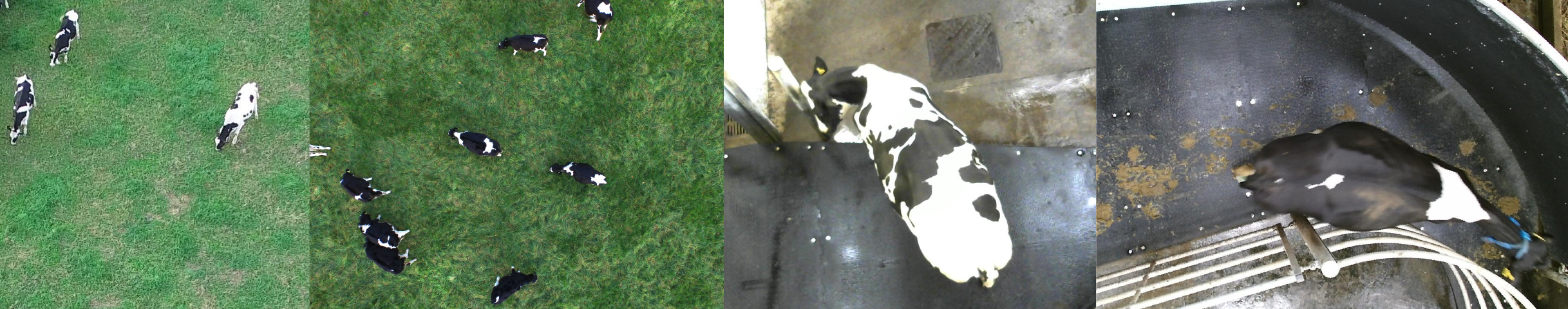}
    \caption{\textbf{Detection \& Localisation Dataset Examples}. Example instances to illustrate the variety in acquisition conditions and environments for the dataset provided for training and testing models performing breed-wide detection and localisation of cattle.}     
    \label{fig:detection-dataset-examples}
\end{figure}

\begin{figure}[htp]
    \centering
    \subfloat{\includegraphics[width=0.49\textwidth]{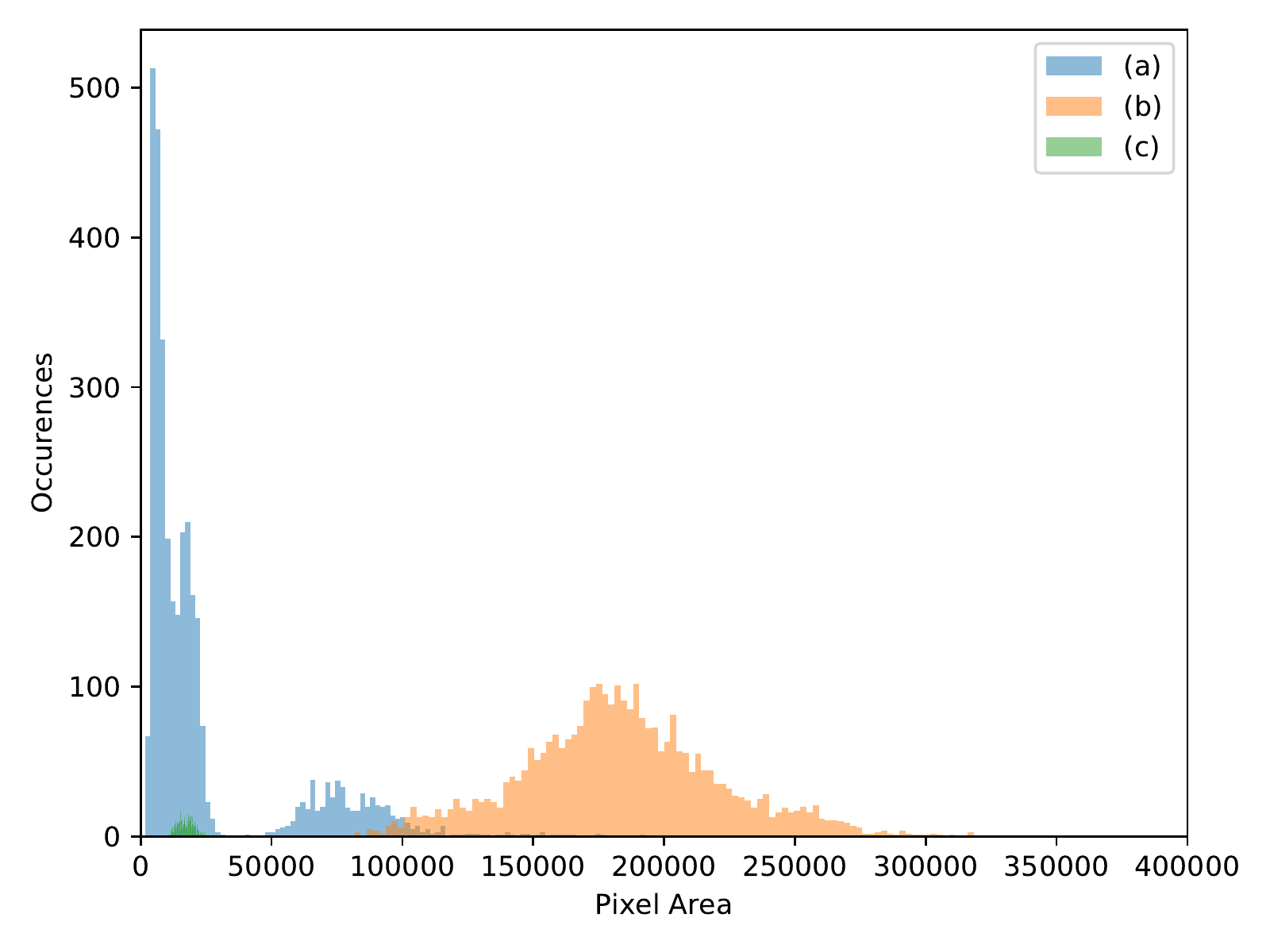}}
    \subfloat{\includegraphics[width=0.49\textwidth]{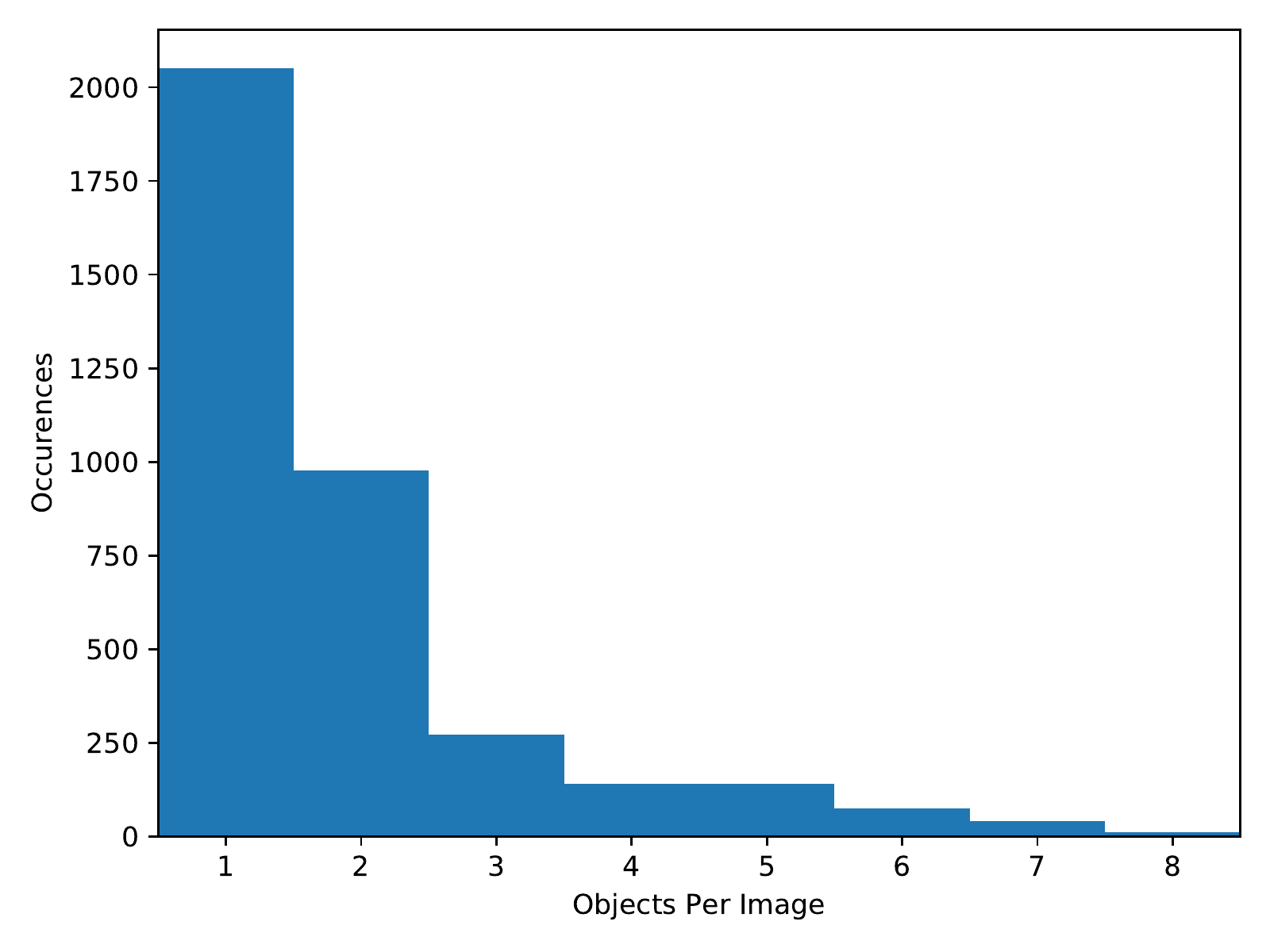}}
    \caption{\textbf{Detection Dataset Distributions}. Distribution of \textit{(left)} annotated object areas in pixels grouped by source:  \textit{(a)} \cite{andrew2019aerial}, \textit{(b)} \cite{andrew2016automatic} and \textit{(c)} \cite{andrew2017visual} and \textit{(right)} object counts per image for the detection component of the OpenCows2020 dataset.}
    \label{fig:detection-distributions}
\end{figure}

The detection and localisation component of the OpenSet2020 dataset consists of whole images with manually annotated cattle regions across in-barn and outdoor settings.
When training a detector on this set, one obtains a model that is widely domain agnostic with respect to the environment, and can be deployed in a variety of farming-relevant conditions.
This component of the dataset consists of a total of $3,707$ images, containing $6,917$ cattle annotations.
For each cow, we manually annotated a bounding box that encloses the animal's torso, excluding the head, neck, legs, and tail in adherence with the VOC 2012 guidelines \cite{pascal-voc-2012}.
This is in order to limit content to a canonical, compact, and minimally deforming species-relevant region.
Illustrative examples from this set are given in Figure~\ref{fig:detection-dataset-examples}.
To facilitate cross validation on this dataset, the images were randomly shuffled and split into 10 folds in a ratio of $8:1:1$ for training, validation and testing, respectively.
The distribution of object sizes and object counts per image are given in Figure~\ref{fig:detection-distributions}, where the difference dataset sources are shown; UAV-acquired footage contains a higher proportion of smaller, distant cattle objects whilst indoor footage generally contains object annotations at a higher resolution.

\subsection{Identification}
\label{subsec:dataset-open-set}

\begin{figure}[h]
    \centering
    \includegraphics[width=0.8\textwidth]{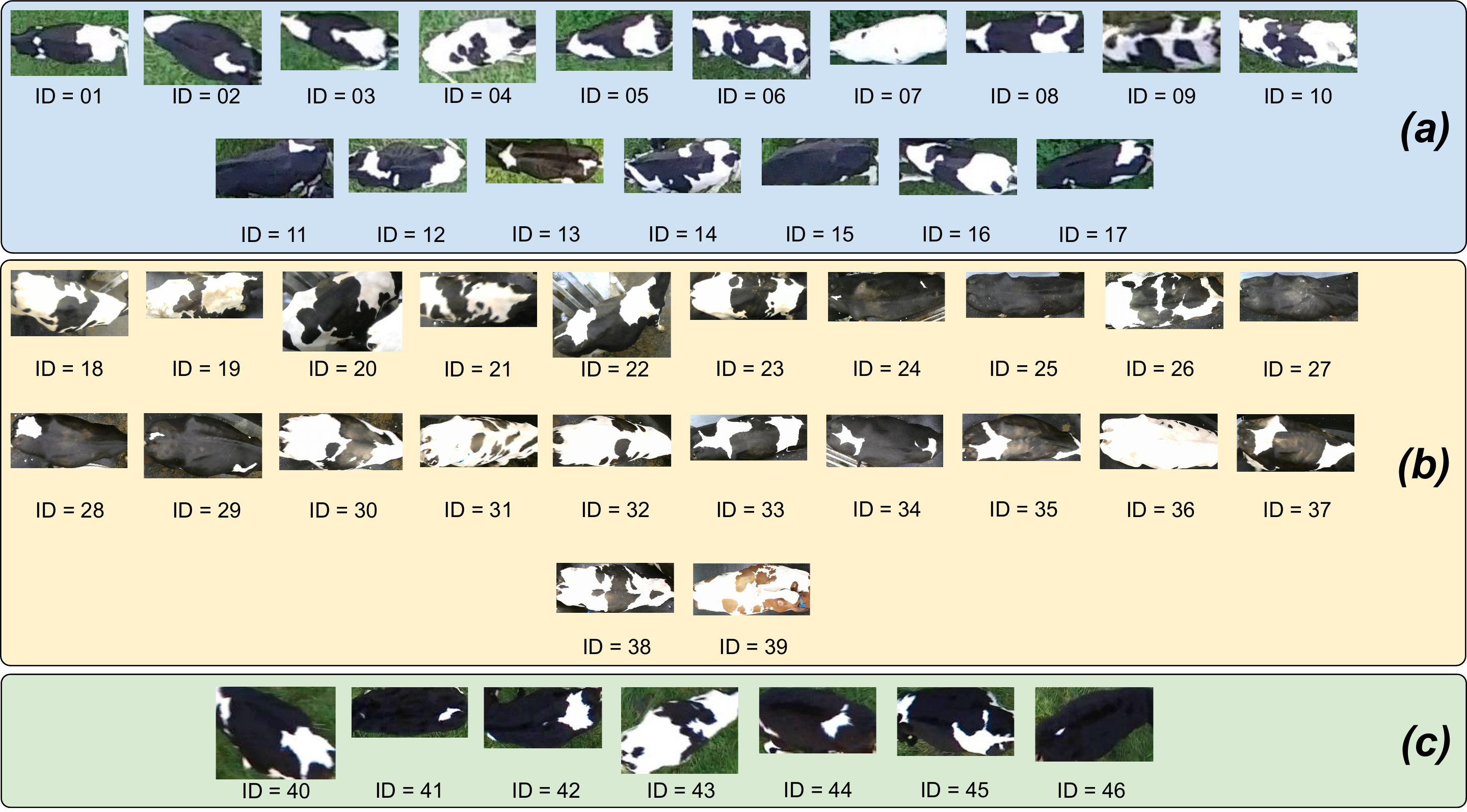}
    \caption{\textbf{Identification Dataset Examples}. Example instances for each of the 46 individuals in the OpenCows2020 dataset grouped by source - \textit{(a)} \cite{andrew2019aerial}, \textit{(b)} \cite{andrew2016automatic} and \textit{(c)} \cite{andrew2017visual}. Observable is the variation in acquisition method, surrounding environment and background, illumination conditions, etc.}
    \label{fig:open-cows-examples}
\end{figure}

\begin{figure}[!htb]
    \centering
    \includegraphics[width=0.85\textwidth]{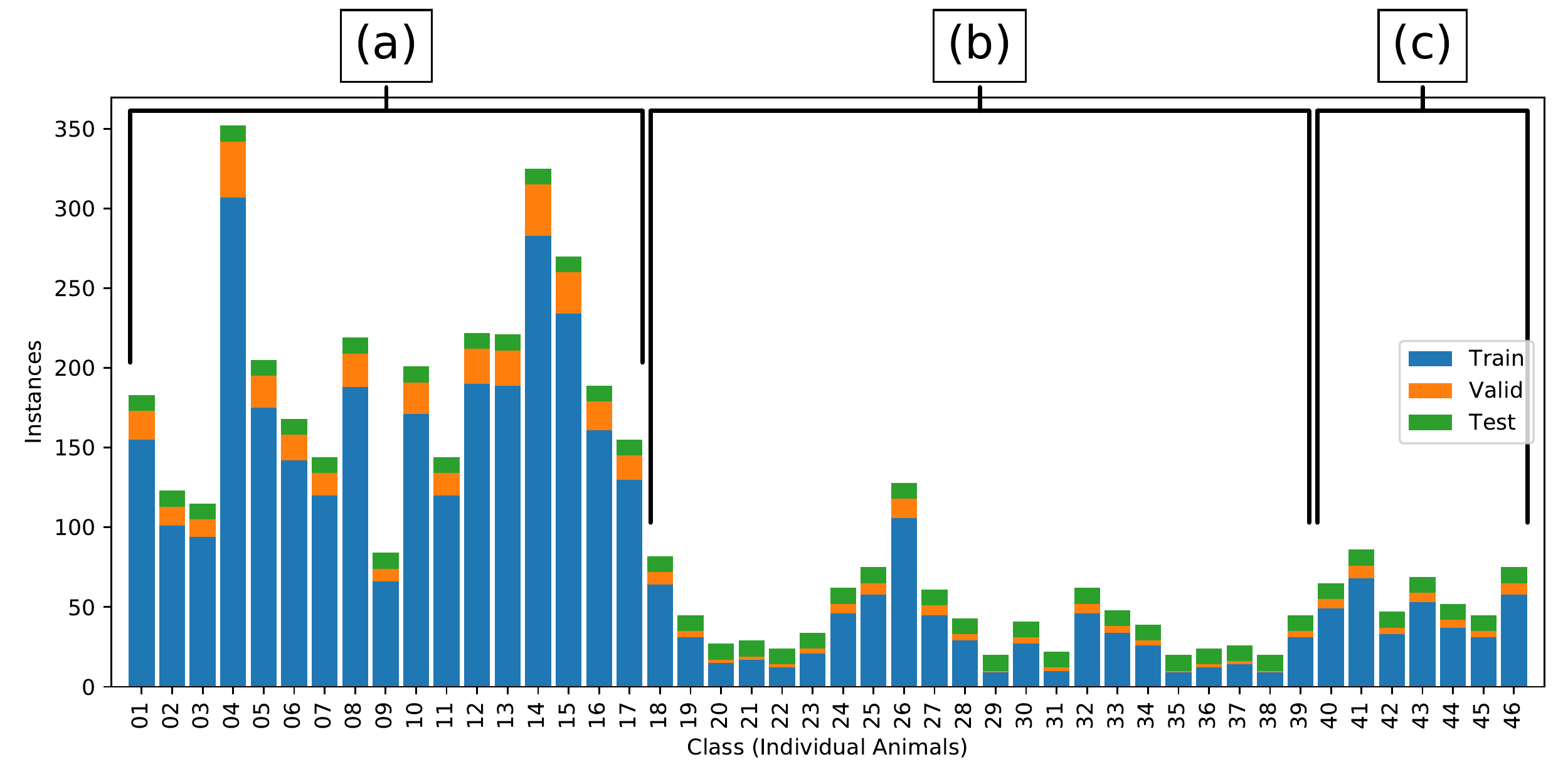}
    \caption{\textbf{Identification Instance Distribution}. The distribution of instances per class for the identification component of the OpenCows2020 dataset. Instances were then randomly split to have exactly 10 testing instances per class whilst those remaining were split into training and validation in a ratio of $9:1$, respectively. Also labelled is the source of each group of categories: \textit{(a)} \cite{andrew2019aerial}, \textit{(b)} \cite{andrew2016automatic} and \textit{(c)} \cite{andrew2017visual}.}
    \label{fig:open-cows-instance-dist}
\end{figure}

The second component of the OpenSet2020 dataset consists of identified cattle from the detection image set.
Individuals with less than 20 instances were discarded in order to have an absolute minimum of 9 training instances, 1 validation instance and 10 testing instances per class (individual animals).
This resulted in a population of $46$ individuals, an average of $103$ instances per class and $4,736$ regions overall.
Instances within a class were randomly split to have exactly 10 testing instances across all individuals. 
The remaining instances were split into training and validation sets in a ratio of $9:1$, respectively.
A random example from each individual is given in Figure~\ref{fig:open-cows-examples} to illustrate the variety in coat patterns, as well as the various acquisition source, method, backgrounds and environments, illumination conditions, etc.
The distribution of instances per class is illustrated in Figure~\ref{fig:open-cows-instance-dist} highlighting the extent of class imbalance that is present across the three sources of data as well as the imbalance within the sources themselves.

\section{Cattle Detection}
\label{sec:detection}

\begin{figure}[htp]
    \centering
    \subfloat{\includegraphics[width=0.49\textwidth]{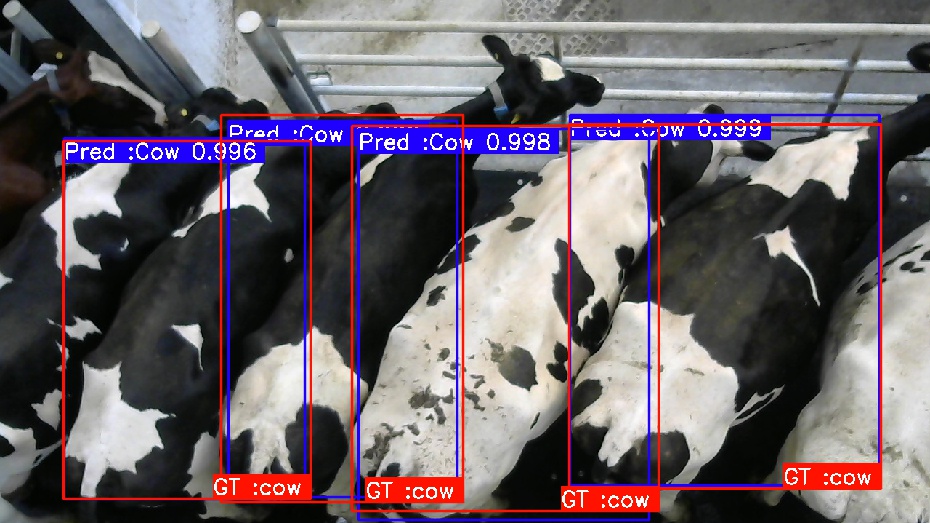}}
    \hfill
    \subfloat{\includegraphics[width=0.49\textwidth]{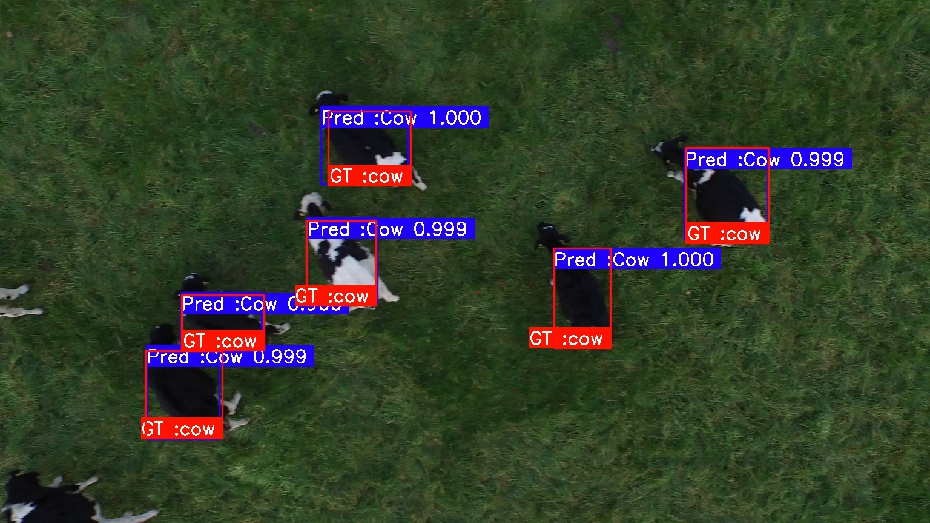}}
    \caption{\textbf{Detection \& Localisation Examples}. Example of \textit{(left)} indoor and \textit{(right)} outdoor instances for object detection. \textit{(red)} Rectangular ground truth bounding boxes for class `cow' and \textit{(blue)} predicted bounding boxes of cow with associated confidence scores. $(x_1,y_1)$ and $(x_2,y_2)$ are the top left and the lower right coordinates of each bounding box, respectively.} 
    \label{fig:detection-example}
\end{figure}

The first stage in the pipeline~(see Fig.~\ref{fig:pipeline-overview}, blue) is the automatic and robust detection and localisation of Holstein-Friesian cattle within relevant imagery.
That is, we want to train a generic breed-wide cattle detector such that for some image input, we receive a set of bounding box coordinates $((x_1,y_1),(x_2,y_2))$ with confidence scores (see Figure \ref{fig:detection-example}) enclosing every cow torso within it as output. 
Note that the object class of (all) cattle is highly diverse, with each individual presenting a different coat pattern. 
The RetinaNet~\cite{lin2017focal}, Faster R-CNN~\cite{ren2015faster}, and YOLOv3~\cite{redmon2018yolov3} architectures were tested and evaluated for this breed recognition task. We will compare their performance in Section \ref{sec:detection-baselines}.


\subsection{Detection Loss}

RetinaNet consists of a backbone feature pyramid network \cite{lin2017feature} followed by two task-specific sub-networks. 
One sub-network performs object classification on the backbone’s output using focal loss, the other regresses the bounding box position.
To implement focal loss, we first define $p_{\mathrm{t}}$ as follows for convenience: 
\begin{equation}
p_{\mathrm{t}}=\left\{\begin{array}{ll}{p} & {\text { if } y=1} \\ {1-p} & {\text { otherwise }}\end{array}\right.
\end{equation}
where $y \in\{\pm 1\}$ is the ground truth and $p$ is the estimated probability when $y$ = 1.
For detection we only need to separate cattle from the background, therefore presenting a binary classification problem.
As such, focal loss is defined as:
\begin{equation}
\label{eq:fl}
\mathrm{FL}=-\alpha_{t}\left(1-p_{t}\right)^{\gamma} \log \left(p_{t}\right)
\end{equation}
where $-\log \left(p_{t}\right)$ is cross entropy for binary classification, $\gamma$ is the modulating factor that balances easy/difficult samples, and $\alpha$ can balance the number of positive/negative samples. 
The focal loss function guarantees that the training process pays attention to positive and difficult samples first.

The regression sub-network predicts four parameters $((P_{x_1},P_{y_1}),(P_{x_2},P_{y_2}))$ representing the offset coordinates $((x_1,y_1),(x_2,y_2))$ between anchor box $A$ and ground-truth box $Y$. 
Their ground-truth offsets $((T_{x_1},T_{y_1}),(T_{x_2},T_{y_2}))$ can be expressed as:
\begin{equation}\begin{aligned}
&T_{x}=\left(Y_{x}-A_{x}\right) / A_{w}\\
&T_{y}=\left(Y_{y}-A_{y}\right) / A_{h}
\end{aligned}\end{equation}
where $Y$ is the ground-truth box and $A$ is the anchor box. 
The width and height of the bounding box are given by $w$ and $h$. 
The regression loss can be defined as:
\begin{equation}
\label{eq:LOC}
\mathbb{L}_{LOC}=\sum_{j \in\{x_1, y_1, x_2, y_2\}} \operatorname{Smooth}_{L 1}\left(P_{j}-Y_{j}\right)
\end{equation}
where Smooth L1 loss is defined as:
\begin{equation} 
\label{eq:sl1}
\operatorname{Smooth}_{L 1}(x)=\left\{\begin{array}{ll}
    0.5 x^{2} & |x|<1 \\
    |x|-0.5 & |x| \geq 1
\end{array}\right.
\end{equation}

Overall, the detection network minimises a combined loss function bringing together Smooth L1 and focal loss components relating to localisation and classification, respectively:
\begin{equation}
    \mathbb{L}_{LOC+FL} = \mathbb{L}_{LOC} + \lambda \cdot \mathbb{L}_{FL},
\end{equation}
where $\mathbb{L}_{LOC}$ and $\mathbb{L}_{FL}$ are defined by equations \ref{eq:LOC} and \ref{eq:fl}, respectively and $\lambda$ is a balancing parameter.


\subsection{Experimental Setup}

Our particular RetinaNet implementation utilises a ResNet-50 backbone~\cite{he2016deep} as the feature pyramid network, with weights pre-trained on ImageNet~\cite{deng2009imagenet}.
Within the detection task, cattle are resolved at a similar size to objects in popular detection datasets. 
As a result, the prior probability of detecting an object ($\pi = 0.01$), anchors and focal loss function parameters are set to the same as in~\cite{lin2017focal}; $\gamma = 2$, $\alpha = 0.25$,  $\lambda = 1$; anchors are implemented in 5 layers (P3 - P7) of feature pyramid network with areas of $32 \times 32$ to $512 \times 512$ for each layer; aspect ratios are $1:2$, $1:1$ and $2:1$.
The network was fine-tuned on the detection component of our dataset using a batch size of $4$ and Stochastic Gradient Descent~\cite{robbins1951stochastic} at an initial learning rate of $1 \times 10^{-5}$ with a momentum of $0.9$~\cite{qian1999momentum} and weight decay at $1 \times 10^{-4}$.
Training time was around 13 hours on an Nvidia V100 GPU (Tesla P100-PCIE-16GB) for $90$ epochs of training.
Additionally, to provide a suitable comparison with other baselines, two popular and seminal architectures -- YOLOv3~\cite{redmon2018yolov3} and Faster R-CNN~\cite{ren2015faster} -- are cross validated (on the same dataset and splits) in the following section.
Final model weights are selected to be those that perform best on the validation set for that fold throughout training (the pocket algorithm \cite{stephen1990perceptron}).
Those weights are then used to evaluate model performance on the respective testing set.


\subsection{Baseline Comparisons and Evaluation}
\label{sec:detection-baselines}

\begin{figure}[h]
    \centering
    \includegraphics[width=0.49\textwidth]{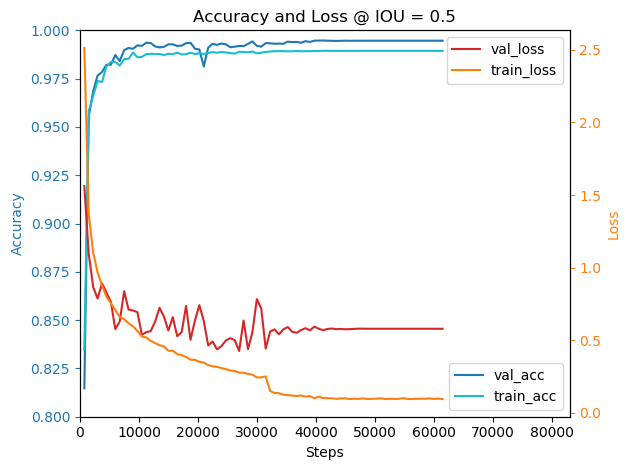}
    \caption{\textbf{RetinaNet Training Graph}. Example of RetinaNet loss and accuracy values during training on the training and validation sets.}
    \label{fig:retinanet-training-graph}
\end{figure}

Quantitative comparisons via 10-fold cross validation of the proposed detection method against classic and recent approaches are shown in Table~\ref{table:detection-baselines}, whilst an example of RetinaNet's response to training is illustrated in Figure \ref{fig:retinanet-training-graph}.
Mean average precision (mAP) is given as the chosen metric to quantitatively compare performance, which for each network is computed via the mean area under the curve for precision-recall across each cross validation fold.
As can be seen in the table, all methods achieve - for all practical purposes - near perfect performance on the detection task, and are therefore all suitable for the application at hand.
Specific parameter choices include a confidence score threshold of $0.5$, non-maximum suppression (NMS) threshold of $0.28$, and the intersection over union (IoU) threshold at $0.5$. 
Whilst the confidence and IoU threshold are commonly used in object detection, we deliberately chose a low NMS threshold, which is justified in the following paragraph.

\begin{table}[]
\resizebox{\textwidth}{!}{%
\begin{tabular}{|l|cc|c|}
\hline
 & \begin{tabular}[c]{@{}c@{}}Pre-trained on\\ COCO \cite{lin2014microsoft}\end{tabular} 
 & \begin{tabular}[c]{@{}c@{}}Pre-trained on\\ ImageNet \cite{deng2009imagenet}\end{tabular} 
 & \begin{tabular}[c]{@{}c@{}}mAP (\%) :\\ $[$minimum, maximum$]$\end{tabular} \\ \hline
YOLO V3 \cite{redmon2018yolov3}  & N & Y & 98.4 : [97.8, 99.2]\\
Faster R-CNN \cite{ren2015faster}(Resnet50 backbone) & Y & N & 99.6 : [98.8, 99.9] \\
RetinaNet \cite{lin2017focal}(Resnet50 backbone) & N & Y & 97.7 : [96.6, 98.8] \\
 \hline
\end{tabular}
}
\caption{\textbf{Quantitative Performance.} Comparative 10-fold cross validated results on the detection component of the OpenCows2020 dataset, where average precision is computed as the area under the curve in precision-recall space for each fold and presented as the mean average precision (mAP) across all folds, as well as the minimum and maximum for each network.}
\label{table:detection-baselines}
\end{table}

\begin{figure}[h]
    \centering
    \subfloat[\label{subfig:detect-fail-1}]{\includegraphics[width=0.325\textwidth]{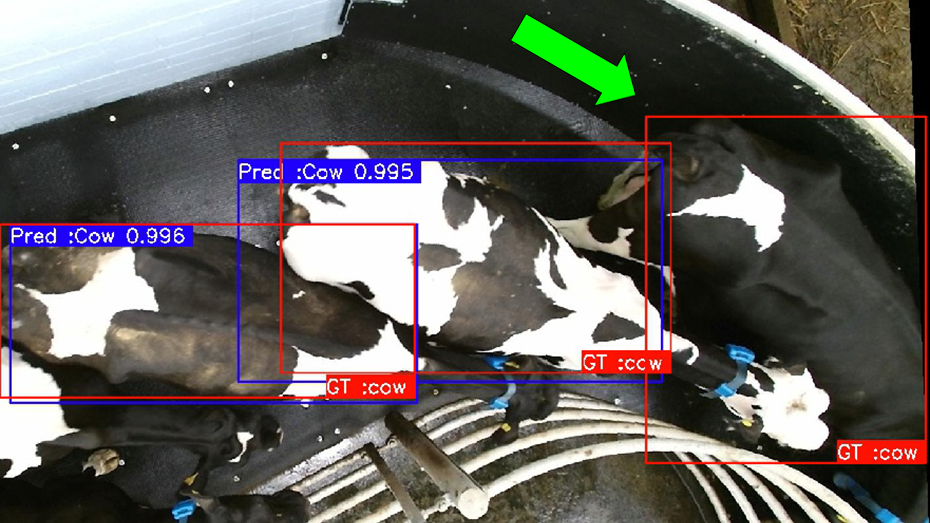}}
    \hfill
    \subfloat[\label{subfig:detect-fail-2}]{\includegraphics[width=0.325\textwidth]{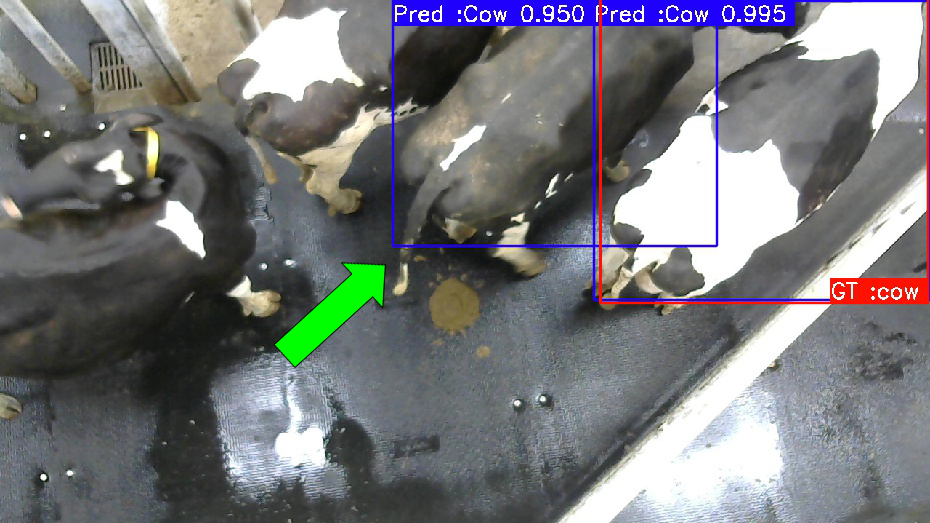}}
    \hfill
    \subfloat[\label{subfig:detect-fail-3}]{\includegraphics[width=0.325\textwidth]{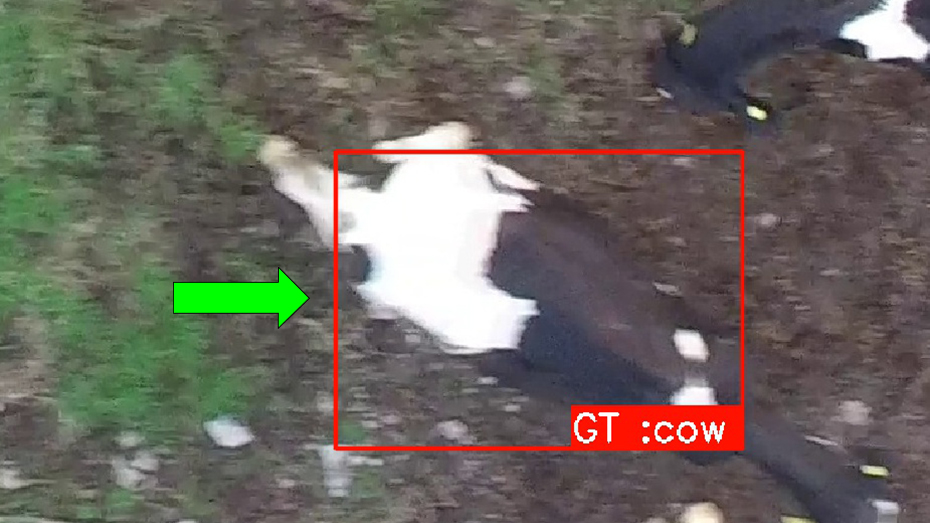}}
    \\
	\subfloat[\label{subfig:detect-fail-4}]{\includegraphics[width=0.325\textwidth]{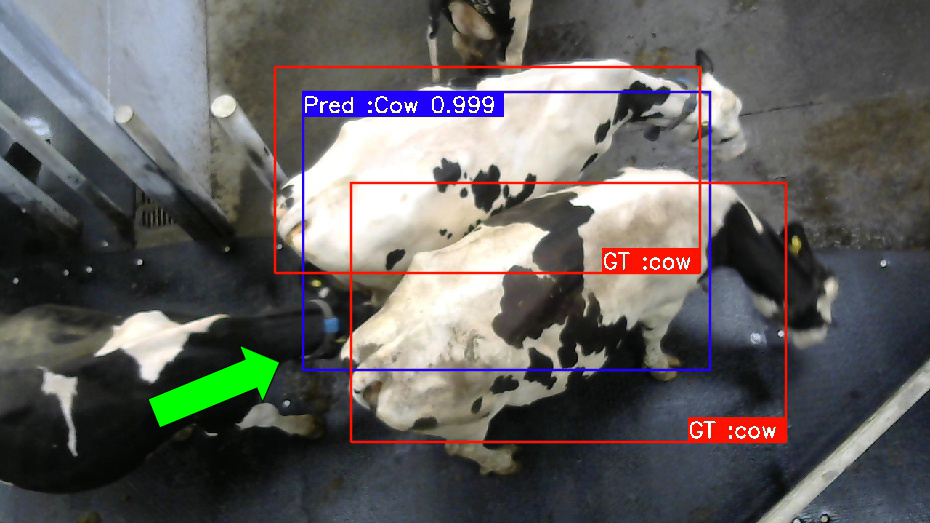}}
    \hfill
    \subfloat[\label{subfig:detect-fail-5}]{\includegraphics[width=0.325\textwidth]{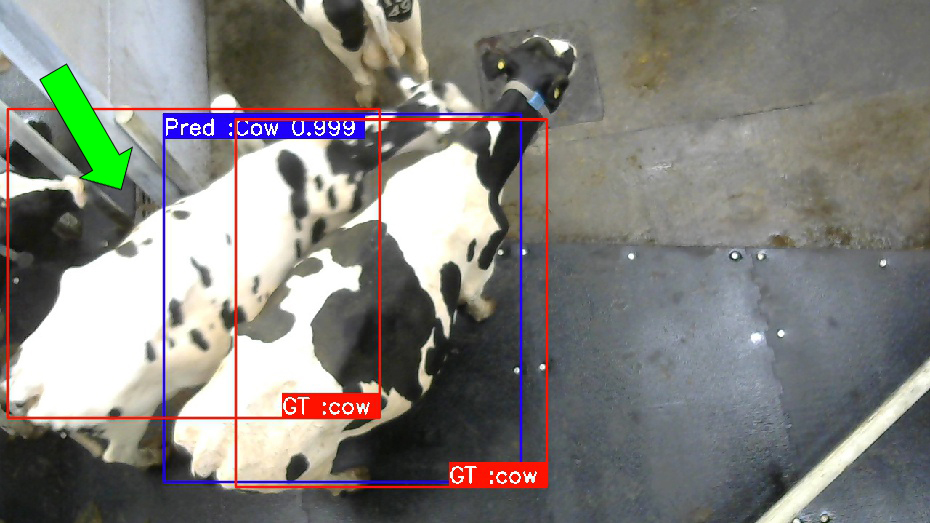}}
    \hfill
    \subfloat[\label{subfig:detect-fail-6}]{\includegraphics[width=0.325\textwidth]{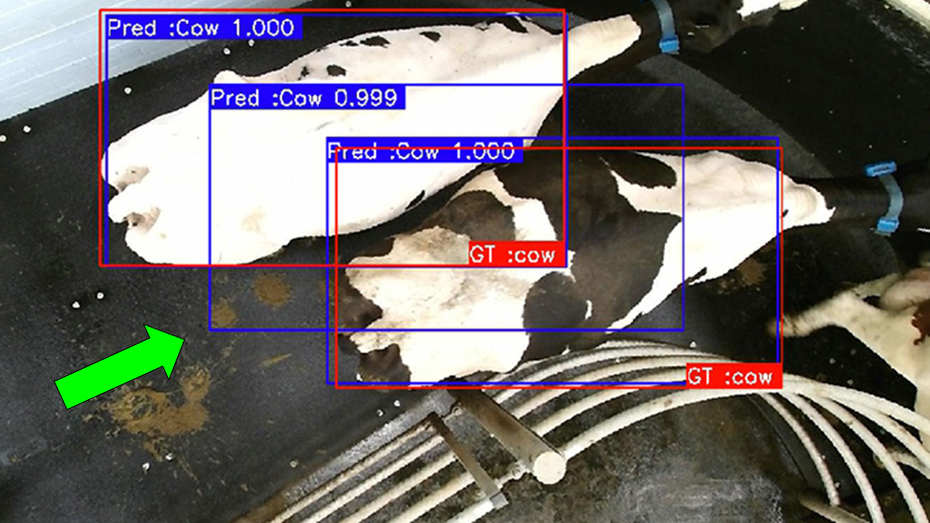}}
    \caption{\textbf{Detection and Localisation Failures of RetinaNet.} Examples cases of rare failures for detecting cattle. (Red): ground truth annotations, (blue): predicted bounding boxes. Examples include \textit{(a)} false negative detection at the boundary of the images, \textit{(b)} false positive detection at the boundary of the images, \textit{(c)} false negative detection, \textit{(d)} inaccurate localisation and false negative detection due to the proximity and alignment of multiple cattle, \textit{(e)} false negative detection due to the proximity and alignment of multiple cattle. \textit{(f)} depicts an example of higher ($0.5$) NMS threshold, where it is not low enough to make a bounding box eliminate its neighbouring high-confidence box.}
    \label{fig:fail-detection}
\end{figure}

Figure \ref{fig:fail-detection} depicts limitations and shows rarely-occurring instances of RetinaNet detection failures.
Examples \ref{subfig:detect-fail-1} and \ref{subfig:detect-fail-2} arise from image boundary clipping following the VOC labelling guidelines \cite{pascal-voc-2012} on object visibility/occlusion which can be avoided in most practical applications by ignoring boundary areas.
In \ref{subfig:detect-fail-4}, the detector gives high detection confidence to the middle part of two adjacent cows, which eliminates two actual prediction boxes through NMS.
In \ref{subfig:detect-fail-5}, false negative detection is the result of closely situated cattle and NMS. 
The closer the cows are, the larger the overlap between their ground truth boxes and corresponding predictions.
In these situations, one predicted box is more likely to be eliminated by NMS.

We chose to keep the NMS threshold as low as possible, otherwise it can occasionally lead to false positive detections in images with groups of crowded cattle.
Figure \ref{subfig:detect-fail-6} depicts this situation with NMS equal to 0.5 rather than the value (0.28) we used for testing.  
When two cattle are standing in close proximity and both have a diagonal heading, a predicted box between the two cows can sometimes be observed. 
This is as a result of one of the intrinsic drawbacks of orthogonal bounding boxes.

\newpage

\section{Open-Set Individual Identification via Metric Learning}
\label{sec:open-ID}

Given robustly identified image regions that contain cattle, we would like to discriminate individuals, seen or unseen, without the costly step of manually labelling new individuals and fully re-training a closed-set classifier. 
The key idea to approach this task is to learn a mapping into a class-distinctive latent space where maps of images of the same individual naturally cluster together. 
Such a feature embedding encodes a latent representation of inputs and, for images, also equates to a significant dimensionality reduction from a matrix $width \times height \times channels$ to an embedding with size $\rm I\!R^n$, where $n$ is the dimensionality of the embedded space. 
In the latent space, distances directly encode input similarity, hence the term of \textit{metric learning}.
To actually classify inputs, after constructing a successful embedding, a lightweight clustering algorithm can be applied to the latent space (e.g. k-Nearest Neighbours) where clusters now represent individuals.


\subsection{Metric Space Building and Loss Functions}
\label{subsec:open-ID-loss-functions}

\begin{figure}[t]
    \centering
    \subfloat[Triplet loss learning objective\label{subfig:triplet-loss}]{\includegraphics[width=0.49\textwidth]{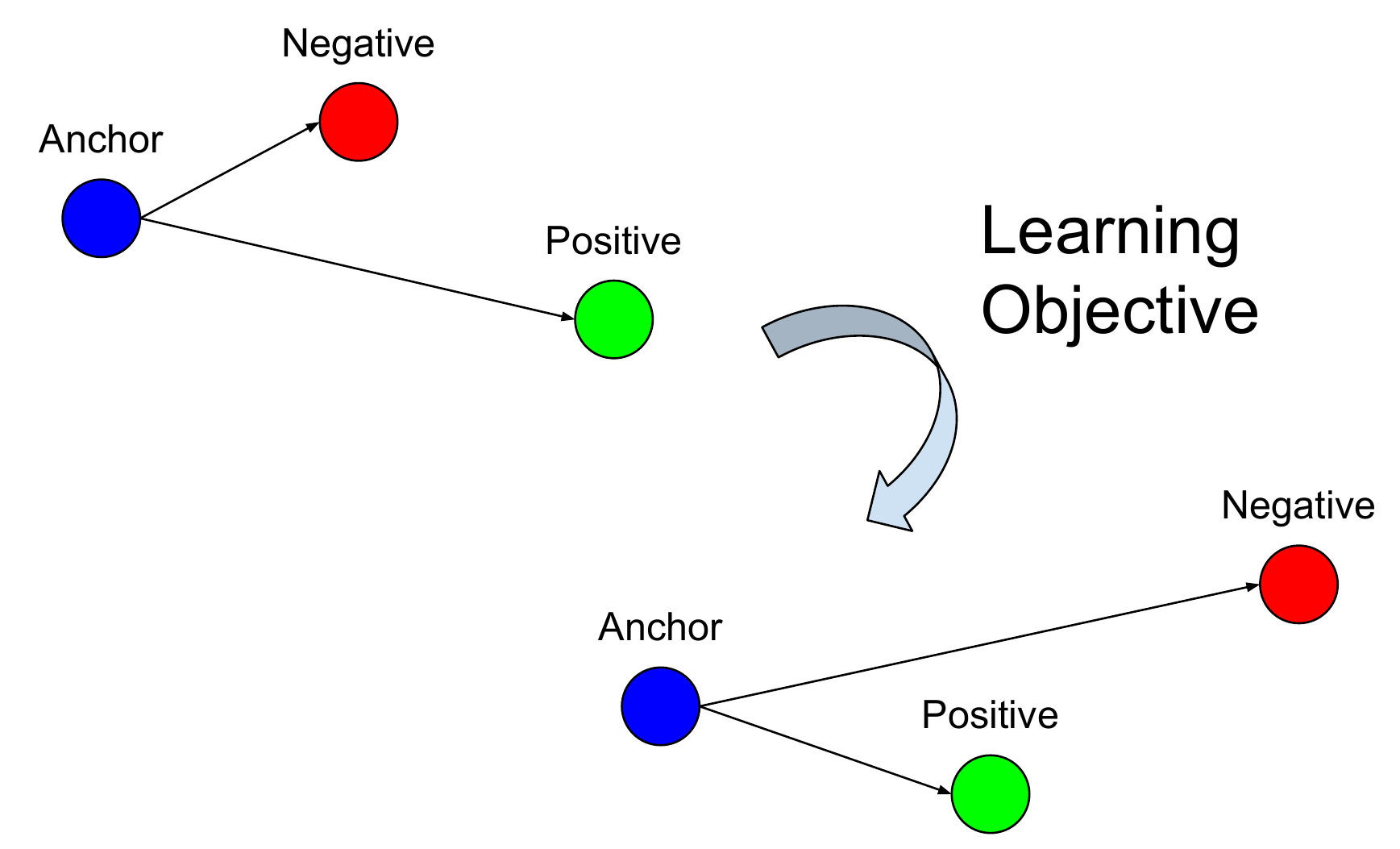}}
    \hfill
    \subfloat[Margin problem\label{subfig:margin-problem}]{\includegraphics[width=0.49\textwidth]{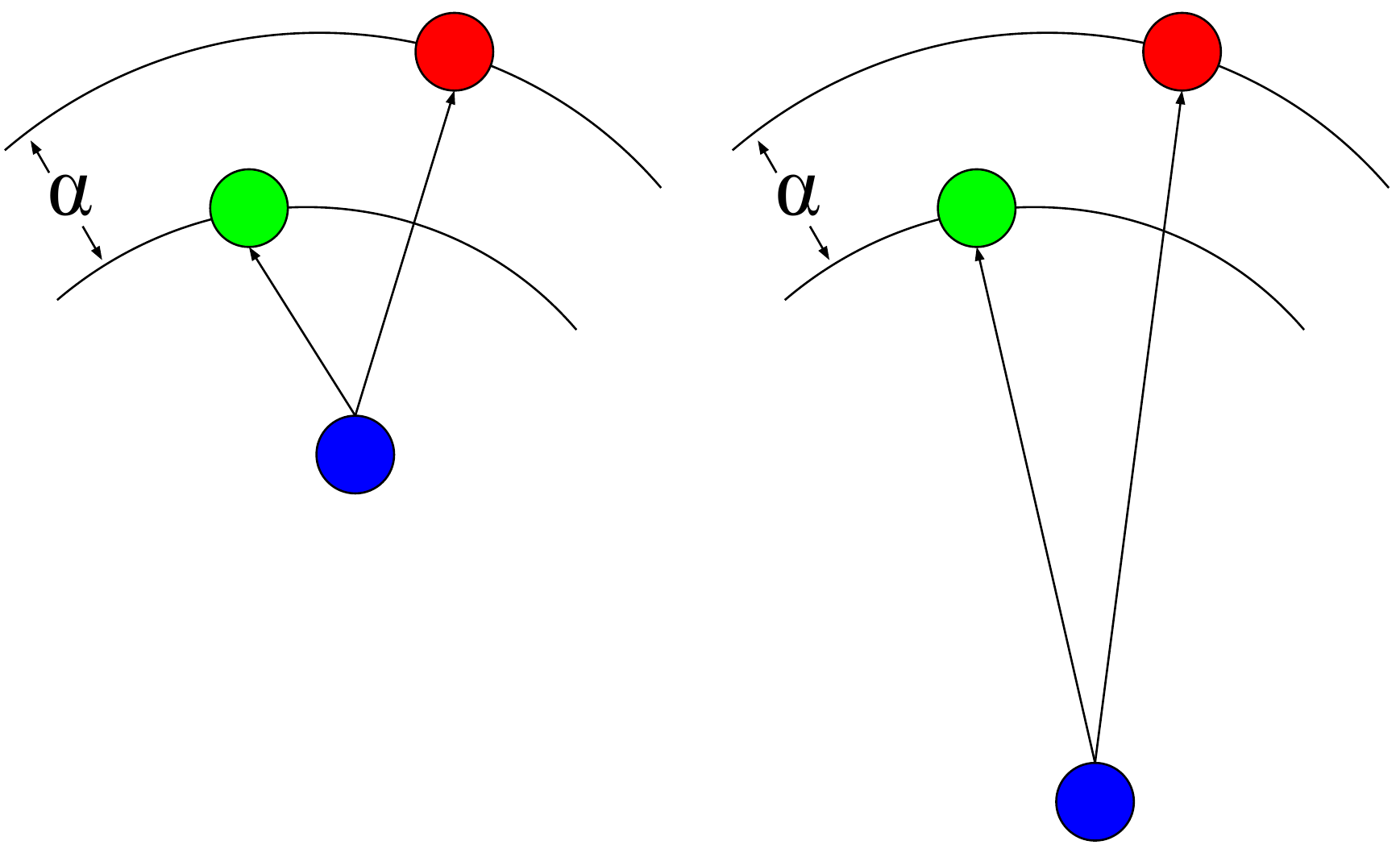}}
    \caption{\textbf{Triplet Loss and the Margin Problem}. \textit{(a)} The triplet loss function aims to minimise the distance between an anchor and a positive instance (both belonging to the same class), whilst maximising the distance between the anchor and a negative (belonging to a different class). However, \textit{(b)} illustrates the problem with the inclusion of a margin $\alpha$ parameter in the triplet loss formulation; it can be satisfied at any distance from the anchor.}
    \label{fig:triplet-loss}
\end{figure}

Success in building this form of latent representation relies heavily -- amongst other factors -- upon the careful choice of a loss function that naturally yields an identity-clustered space.
A seminal example in metric learning originates from the use of Siamese architectures~\cite{hadsell2006dimensionality}, where image pairs $X_1,X_2$ are passed through a dual stream network with coupled weights to obtain their embedding. 
Weights $\theta$ are shared between two identical network streams $f_\theta$:
\begin{equation}
\begin{split}
    x_1&=f_\theta(X_1),\\
    x_2&=f_\theta(X_2).
\end{split}
\end{equation}
The authors proposed training this architecture with a contrastive loss to cluster instances according to their class:
\begin{equation}
    \mathbb{L}_{Contrastive} = (1-Y)\frac{1}{2}d(x_1,x_2)+Y\frac{1}{2}max(0,\alpha-d(x_1,x_2)),
\end{equation}
where $Y$ is a binary label denoting similarity or dissimilarity on the inputs $(X_1,X_2)$, and $d(\cdot,\cdot)$ is the Euclidean distance between two embeddings with dimensionality $n$.
The problem with this formulation is that it cannot \textit{simultaneously} encourage learning of visual similarities and dissimilarities, both of which are critical for obtaining clean, well-separated clusters on our coat pattern differentiation task.
This shortcoming can be overcome by a triplet loss formulation~\cite{schroff2015facenet}; utilising the embeddings $x_a,x_p,x_n$ of a triplet containing three image inputs $(X_a,X_p,X_n)$ denoting an anchor, a positive example from the same class, and a negative example from a different class, respectively.
The idea being to encourage minimal distance between the anchor $x_a$ and the positive $x_p$, and maximal distance between the anchor $x_a$ and the negative sample $x_n$ in the embedded space.
Figure~\ref{subfig:triplet-loss} illustrates the learning goal, whilst the loss function is given by:
\begin{equation}
\label{eq:standard-triplet-loss}
    \mathbb{L}_{TL} = \max(0,d(x_a, x_p) - d(x_a, x_n) + \alpha),
\end{equation}
where $\alpha$ denotes a constant margin hyperparameter.
The inclusion of the constant $\alpha$ often turns out to cause training issues since the margin can be satisfied at any distance from the anchor; Figure \ref{subfig:margin-problem} illustrates this problem.
a recent formulation named reciprocal triplet loss alleviates this limitation \cite{masullo2019goes}, which removes the margin hyperparameter altogether:
\begin{equation}
\label{eq:reciprocal-triplet-loss}
    \mathbb{L}_{RTL} = d(x_a, x_p) + \frac{1}{d(x_a, x_n)}.
\end{equation}

Recent work~\cite{lagunes2019learning} has demonstrated improvements in open-set recognition on various datasets~\cite{hodan2017t, wang2017object} via the inclusion of a SoftMax term in the triplet loss formulation during training given by:
\begin{equation}
    \mathbb{L}_{SoftMax+TL} = \mathbb{L}_{SoftMax} + \lambda \cdot \mathbb{L}_{TL},
\end{equation}
where
\begin{equation}
\label{eq:softmax-loss}
    \mathbb{L}_{SoftMax} = -log\left(\frac{e^{x_{class}}} {\sum_ie^{x_i}}\right),
\end{equation}
and where $\lambda$ is a constant weighting hyperparameter and $\mathbb{L}_{TL}$ is standard triplet loss as defined in equation \ref{eq:standard-triplet-loss}.
For our experiments, we select~$\lambda=0.01$ as suggested in the original paper~\cite{lagunes2019learning} as the result of a parameter grid search.
This formulation is able to outperform the standard triplet loss approach since it combines the best of both worlds -- fully supervised learning and a separable embedded space.
Most importantly for the task at hand, we propose to combine a fully supervised loss term as given by Softmax loss with the reciprocal triplet loss formulation which removes the necessity of specifying a margin parameter.
This combination is novel and given by:
\begin{equation}
    \mathbb{L}_{SoftMax+RTL} = \mathbb{L}_{SoftMax} + \lambda \cdot \mathbb{L}_{RTL},
\end{equation}
where $\mathbb{L}_{SoftMax}$ and $\mathbb{L}_{RTL}$ are defined by equations~\ref{eq:reciprocal-triplet-loss} and~\ref{eq:softmax-loss} above, respectively.
Comparative results for all of these loss functions are given in our experiments as follows.


\section{Experiments}
\label{sec:experiments}

In the following section, we compare and contrast different triplet loss functions to quantitatively and qualitatively show performance differences on our task of open-set identification of Holstein-Friesian cattle.
The goal of the experiments carried out here is to investigate the extent to which different feature embedding spaces are suitable for our specific open-set classification task.
Within the context of the overall identification pipeline given in Figure~\ref{fig:pipeline-overview}, we will assume that the earlier stage (as described in Section \ref{sec:detection}) has successfully detected the presence of cattle and extracted good-quality regions of interest.
These regions are now ready to be identified, as assessed in these experiments.


\subsection{Experimental Setup}
\label{subsec:exp-setup-open-set}

The employed embedding network utilises a ResNet50 backbone~\cite{he2016deep}, with weights pre-trained on ImageNet~\cite{deng2009imagenet}. 
The final fully connected layer was set to have $n=128$ outputs, defining the dimensionality of the embedding space. 
This dimensionality choice was founded on existing research suggesting $n=128$ to be suitable for fine-grained recognition tasks such as face recognition~\cite{schroff2015facenet} and image class retrieval~\cite{balntas2016learning}.
In each experiment, the network was fine-tuned on the training portion of the identification regions in the OpenCows2020 dataset over $100$ epochs with a batch size of $16$.
Final model weights were selected to be those with highest performance on the validation set throughout training (the pocket algorithm \cite{stephen1990perceptron}) to cope with potential overfitting.
These final model weights were then used to obtain classification performance on the testing set.
We chose Stochastic Gradient Descent~\cite{robbins1951stochastic} as the optimiser, set to an initial learning rate of $1 \times 10^{-5}$ with momentum $0.9$~\cite{qian1999momentum} and weight decay $1 \times 10^{-4}$.
Of note is that we found the momentum component led to significant instability during training with reciprocal triplet loss, thus we disabled it for runs using that function.
Finally, for a comparative closed-set classifier chosen as another baseline, the same ResNet50 architecture was used.

\begin{figure}[h]
    \centering
    \includegraphics[width=0.7\textwidth]{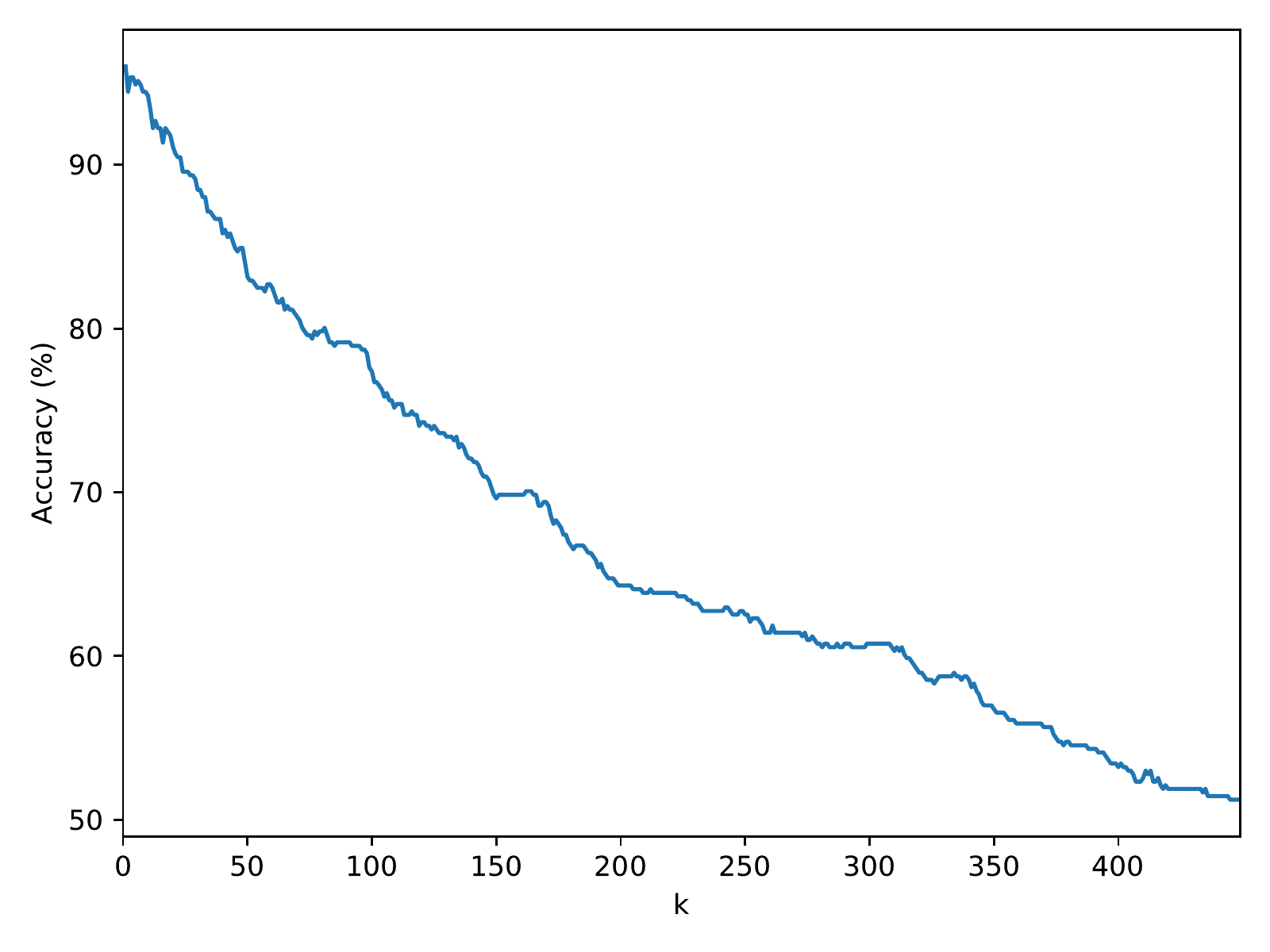}
    \caption{\textbf{$\mathbf{k}$-NN Parameter Search}. The accuracy achieved when classifying the latent representation of the validation set for varying choices of $k$ nearest neighbours. This search was performed on a 50\% open problem.}
    \label{fig:knn-grid-search}
\end{figure}

Once an image is passed through the network, we obtain its $n$-dimensional embedding $x$.
We then used $k$-NN for classification with $k=5$ as suggested by similar research~\cite{lagunes2019learning} and as confirmed via a parameter search (illustrated in Fig. \ref{fig:knn-grid-search}).
Using $k$-NN to classify unseen classes operates by projecting every non-testing instance from every class into the latent space; both those seen and unseen during the network training.
Subsequently, every testing instance (of known and unknown individuals) is also projected into the latent space.
Finally, each testing instance is classified from votes from the surrounding $k$ nearest embeddings from non-testing instances.
Accuracy is then defined as the number of correct predictions divided by the cardinality of the testing set.

To validate the model in its capacity to generalise from seen to unseen individuals, we vary the ratio of classes that are withheld entirely from training, forming the unknown set, with $r=\{0.1, 0.17, 0.25, 0.33, 0.5, 0.6, 0.7, 0.8, 0.9\}$.
Unknown classes are chosen in adherence with the respective ratio by randomly sampling equally from each image source (indoor or outdoor, see Section \ref{subsec:dataset-open-set}), so as to avoid a bias towards the background or object resolution.
At each ratio, $n=10$ repetitions are randomly generated in this way in order to establish a more accurate picture of model performance at that ratio of known versus unknown classes.
By varying this proportion, we investigate the model's performance on an increasingly open problem.
Importantly, these $|r| \times n$ splits remain constant throughout experimentation to ensure consistency and enable quantitative comparison between the implemented training strategies.


\subsubsection{Identity Space Mining Strategies}

During training, one observes the network learning quickly and, as a result, a large fraction of triplets are rendered relatively uninformative.
The commonly-employed remedy is to mine triplets \textit{a priori} for difficult examples.
This offline process was superseded by Hermans \textit{et al.} in their 2017 paper~\cite{hermans2017defense} who proposed two \textit{online} methods for mining more appropriate triplets, `batch hard' and `batch all'.
Triplets are mined within each mini-batch during training and their triplet loss computed over the selections.
In this way, a costly offline search before training is no longer necessary.
Consequently, we employ `batch hard' here as our online mining strategy, as given by:
\begin{equation}
    \mathbb{L}_{BH}(X) = 
    \overbrace{\sum_{i=1}^P \sum_{a=1}^K}^\text{all anchors}
    \max\left(0, 
    \overbrace{\max_{p=1...K}d(x_a^i,x_p^i)}^\text{hardest positive} - 
    \overbrace{\min_{\substack{j=1...P; \\ n=1..K; \\ j\neq i}} d(x_a^i,x_n^j)}^\text{hardest negative} + 
    \alpha \right),
\end{equation}
where $X$ is the mini-batch of triplets, $P$ are the anchor classes and $K$ are the images for those anchors.
This formulation selects moderate triplets overall, since they are the hardest examples within each mini-batch, which is in turn a small subset of the training data.
We use this mining strategy for all of the tested loss functions given in the following results section.


\subsection{Results}
\label{subsec:findings}

\begin{figure}[!htb]
    \centering
    \includegraphics[width=1.0\textwidth]{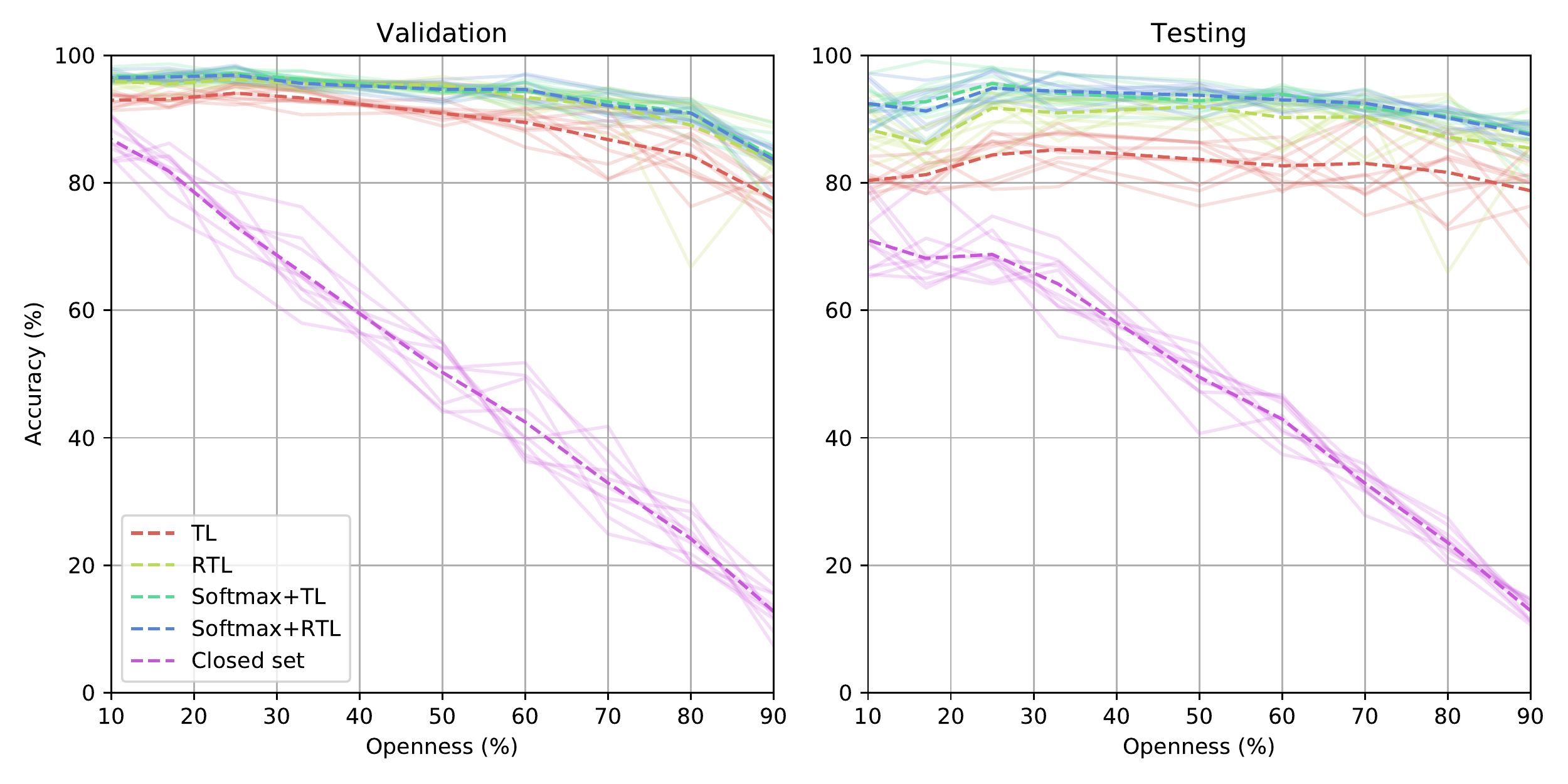}
    \caption{\textbf{Open-Set Generalisation Ability.} Average, minimum, and maximum accuracy across folds versus how open the problem is; that is, the proportion of all identity classes that are withheld entirely during training. Plotted are the differing responses based on the employed loss function where TL and RTL denote standard triplet loss and reciprocal triplet loss respectively, and ``SoftMax +'' denotes a weighted mixture of cross-entropy and triplet loss functions as suggested by \cite{lagunes2019learning}. Also included is a baseline to highlight the unsuitability of a traditional closed-set classifier.}
    \label{fig:acc-vs-open}
\end{figure}

\begin{figure}[h]
    \centering
    \includegraphics[width=0.6\textwidth]{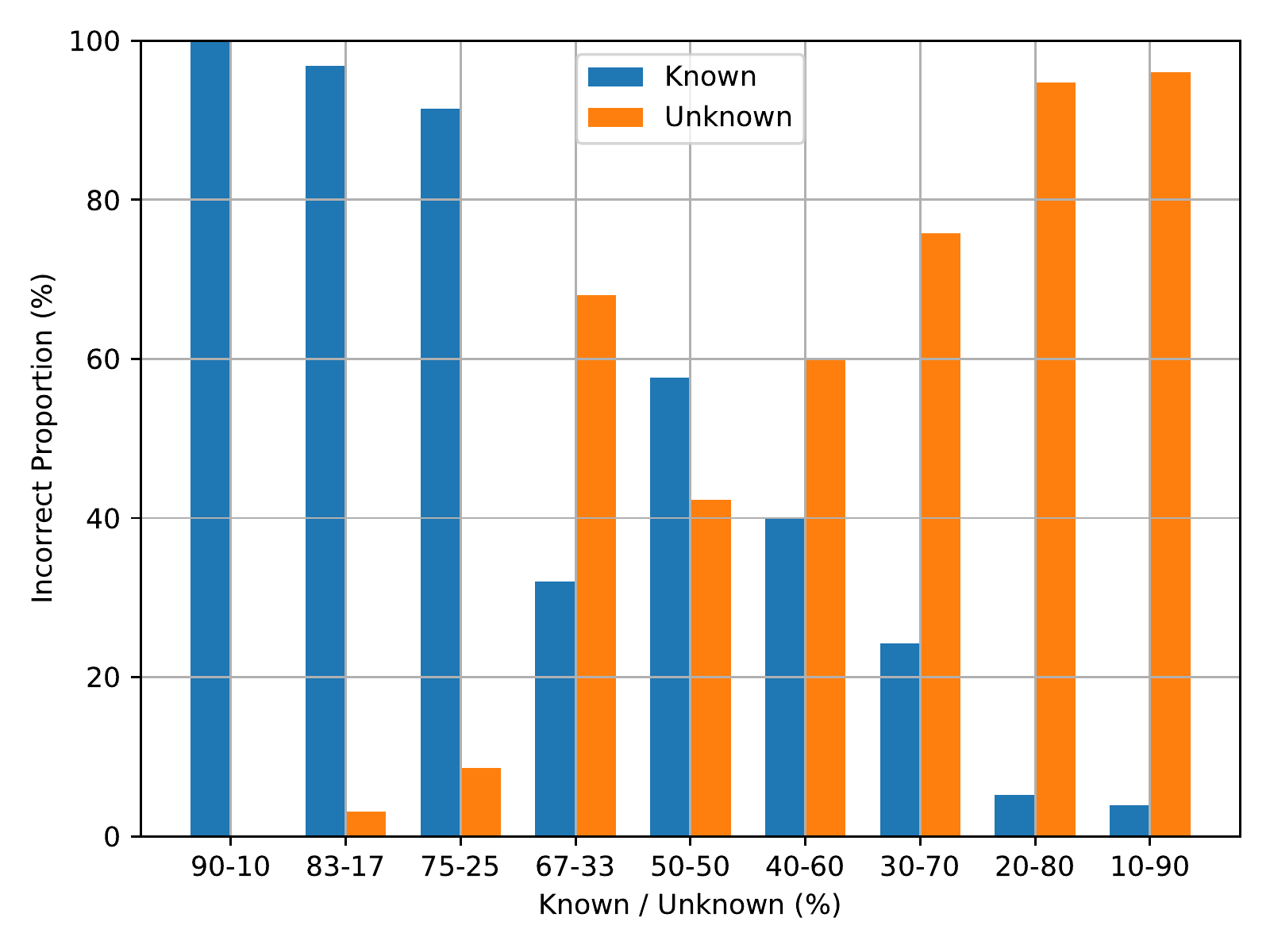}
    \caption{\textbf{Error Proportion vs. Openness.} Where the proportion of error lies (in the known or unknown set) versus how open-set the problem is. Values were calculated from embeddings trained via Softmax and reciprocal triplet loss. These results were found to be consistent across all employed loss functions.}
    \label{fig:error-proportion}
\end{figure}

Key quantitative results for our experiments are given in Table~\ref{table:open-cows-results}, with Figure~\ref{fig:acc-vs-open} illustrating the ability for the implemented methods to cope with an increasingly open-set problem compared to a standard closed-set CNN-based classification baseline using Softmax and cross-entropy loss.
As one would expect, the closed-set method has a linear relationship with how open the identification problem is set (horizontal axis); the baseline method can in no way generalise to unseen classes by design.
In stark contrast, all embedding-based methods can be seen to substantially outperform the implemented baseline, demonstrating their suitability to the problem at hand.
Encouragingly, as shown in Figure \ref{fig:error-proportion}, we found that identification error had no tendency to originate from the unknown identity set.
In addition, we found that there was no tendency for the networks to find any one particular image source (e.g. indoors, outdoors) to be more difficult, and instead found that a slightly higher proportion of error tended to reside within animals/categories with fewer instances, as one would expect.

\clearpage
\begin{landscape}
\begin{table}[b]
\resizebox{1.8\textwidth}{!}{%
\begin{tabular}{|c|l|ccccccccc|}
\hline
& & \multicolumn{9}{c|}{Average Accuracy (\%) : [Minimum, Maximum]} \\ \cline{2-11}
& Known / Unknown (\%)  & 90 / 10  & 83 / 17 & 75 / 25 & 67 / 33 & 50 / 50 & 40 / 60 & 30 / 70 & 20 / 80 & 10 / 90 \\
\hline
\multirow{5}{*}{\textit{Validation}} & Cross-entropy (Closed-set) & $86.82:[83.33,90.67]$ & $81.80:[74.67,86.22]$ & $73.16:[65.33,78.67]$ & $65.96:[58.00,76.22]$ & $50.27:[44.00,54.89]$ & $42.47:[36.22,51.78]$ & $32.89:[24.89,41.78]$ & $24.16:[20.00,29.78]$ & $12.73:[7.33,16.89]$
 \\
& Triplet Loss \cite{schroff2015facenet} & $92.97:[91.35,94.46]$ & $93.13:[91.80,95.34]$ & $94.08:[92.24,95.57]$ & $93.33:[90.69,94.90]$ & $90.91:[88.91,93.13]$ & $89.48:[85.59,91.80]$ & $86.78:[80.49,92.02]$ & $84.26:[76.27,88.69]$ & $77.49:[72.06,82.93]$ \\
& Reciprocal Triplet Loss \cite{masullo2019goes} & $95.94:[94.68,97.34]$ & $95.52:[92.90,97.78]$ & $96.27:[94.68,97.34]$ & $95.52:[94.46,96.45]$ & $95.32:[94.01,96.67]$ & $93.44:[90.69,95.34]$ & $92.11:[86.92,94.90]$ & $89.05:[66.74,93.35]$ & $83.55:[76.72,89.58]$ \\
& Softmax + Triplet Loss \cite{lagunes2019learning} & $96.74:[95.79,98.23]$ & $96.70:[95.12,98.67]$ & $97.14:[95.79,98.23]$ & $96.23:[94.24,97.56]$ & $94.41:[92.68,95.12]$ & $94.37:[91.13,96.23]$ & $92.71:[88.91,94.90]$ & $91.00:[86.92,92.90]$ & $84.10:[76.27,89.36]$ \\
& \begin{tabular}[l]{@{}l@{}}\textbf{(Ours)} \textit{Softmax +}\\\textit{Reciprocal Triplet Loss}\end{tabular} & $96.54:[95.79,97.78]$ & $96.61:[95.79,98.00]$ & $96.90:[95.57,98.45]$ & $95.61:[94.90,96.23]$ & $94.68:[92.46,96.23]$ & $94.66:[92.46,97.12]$ & $92.13:[89.58,94.68]$ & $90.95:[89.14,93.13]$ & $83.64:[78.27,85.81]$ \\ \hline

\multirow{5}{*}{\textit{\textbf{Testing}}} & Cross-entropy (Closed-set) & $71.02:[65.22,79.78]$ & $68.15:[63.48,80.43]$ & $68.76:[64.13,74.78]$ & $64.11:[55.87,71.30]$ & $49.52:[40.65,54.78]$ & $42.91:[37.39,46.74]$ & $32.87:[27.83,35.87]$ & $23.54:[20.22,27.39]$ & $12.87:[10.65,14.78]$ \\
& Triplet Loss \cite{schroff2015facenet} & $80.37:[77.01,84.16]$ & $81.28:[78.31,84.82]$ & $84.38:[78.96,88.07]$ & $85.23:[79.39,89.37]$ & $83.67:[76.36,90.46]$ & $82.67:[78.52,87.20]$ & $83.06:[74.84,90.46]$ & $81.65:[72.67,88.29]$ & $78.76:[67.03,85.47]$ \\
& Reciprocal Triplet Loss \cite{masullo2019goes} & $88.37:[82.65,93.49]$ & $86.18:[81.78,94.58]$ & $91.74:[84.38,95.88]$ & $90.98:[86.55,93.49]$ & $92.00:[88.29,95.66]$ & $90.26:[84.82,93.06]$ & $90.33:[83.08,94.36]$ & $87.18:[65.94,93.93]$ & $85.44:[78.09,91.76]$ \\
& Softmax + Triplet Loss \cite{lagunes2019learning} & $92.15:[87.85,97.18]$ & $\mathbf{92.73:[86.77,99.13]}$ & $\mathbf{95.60:[92.62,98.26]}$ & $94.06:[90.46,97.18]$ & $92.89:[90.02,96.10]$ & $\mathbf{93.88:[91.54,95.44]}$ & $91.80:[88.72,94.79]$ & $\mathbf{90.46:[86.55,92.84]}$ & $\mathbf{87.79:[80.48,91.11]}$ \\
& \begin{tabular}[l]{@{}l@{}}\textbf{(Ours)} \textit{Softmax +}\\\textit{Reciprocal Triplet Loss}\end{tabular} & $\mathbf{92.45:[88.07,97.18]}$ & $91.26:[86.12,96.10]$ & $94.84:[92.41,97.83]$ & $\mathbf{94.36:[91.32,97.40]}$ & $\mathbf{93.75:[90.24,95.66]}$ & $93.02:[90.89,94.58]$ & $\mathbf{92.49:[90.46,94.58]}$ & $90.22:[87.64,91.97]$ & $87.55:[83.51,89.80]$ \\ \hline
\end{tabular}
}
\caption{\textbf{Average Accuracies.} Average, minimum, and maximum accuracies from $n=10$ repetitions for varying ratios of known to unknown classes within the OpenCows2020 dataset consisting of 46 individuals. These results are also illustrated in Figure \ref{fig:acc-vs-open}.}
\label{table:open-cows-results}
\end{table}
\end{landscape}
\clearpage

One issue we encountered is that when there are only a small number of training classes, the model can quickly learn to satisfy that limited set; achieving near-zero loss and 100\% accuracy on the validation data for those seen classes.
However, the construction of the latent space is widely incomplete and there is no room for the model to learn any further, and thus performance on novel classes cannot be improved.
For best performance in practice, we therefore suggest to utilise as wide an identity landscape as possible (many individuals) to carve out a diverse latent space capturing a wide range of intra-breed variance.
The avoidance of overfitting is crucial where eventual perfect performance on a small set of known validation identities does not allow performance to generalise to novel classes.
The reciprocal triplet loss formulation performs slightly better across the learning task, which is reflected quantitatively in our findings (see Figure \ref{fig:acc-vs-open}).
Thus, we suggest utilisation of RTL over the original triplet loss function for the task at hand.

\begin{figure}[h]
    \centering
    \subfloat[50\% open]{\includegraphics[width=0.49\textwidth]{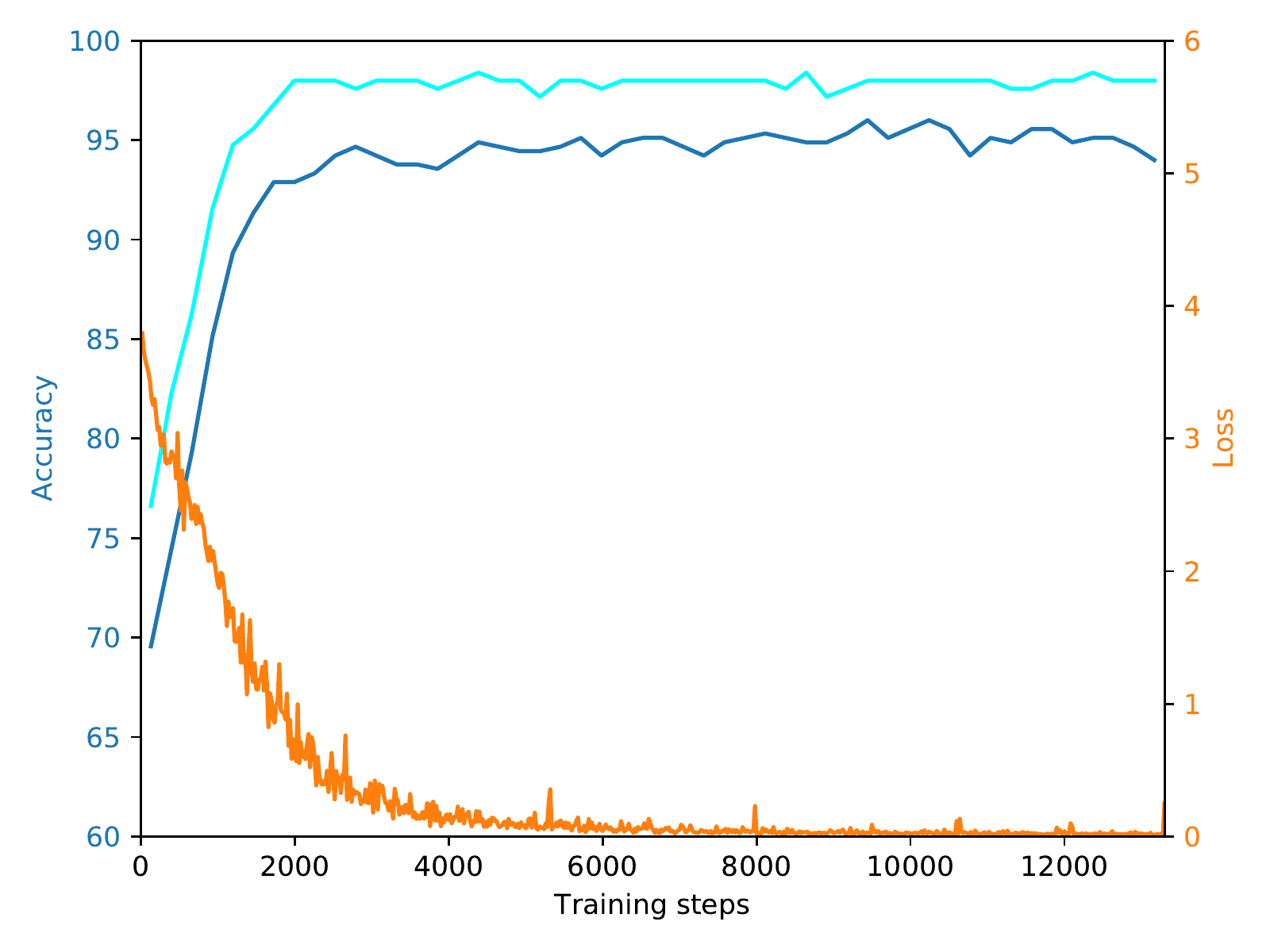}}
    \hfill
    \subfloat[10\% open]{\includegraphics[width=0.49\textwidth]{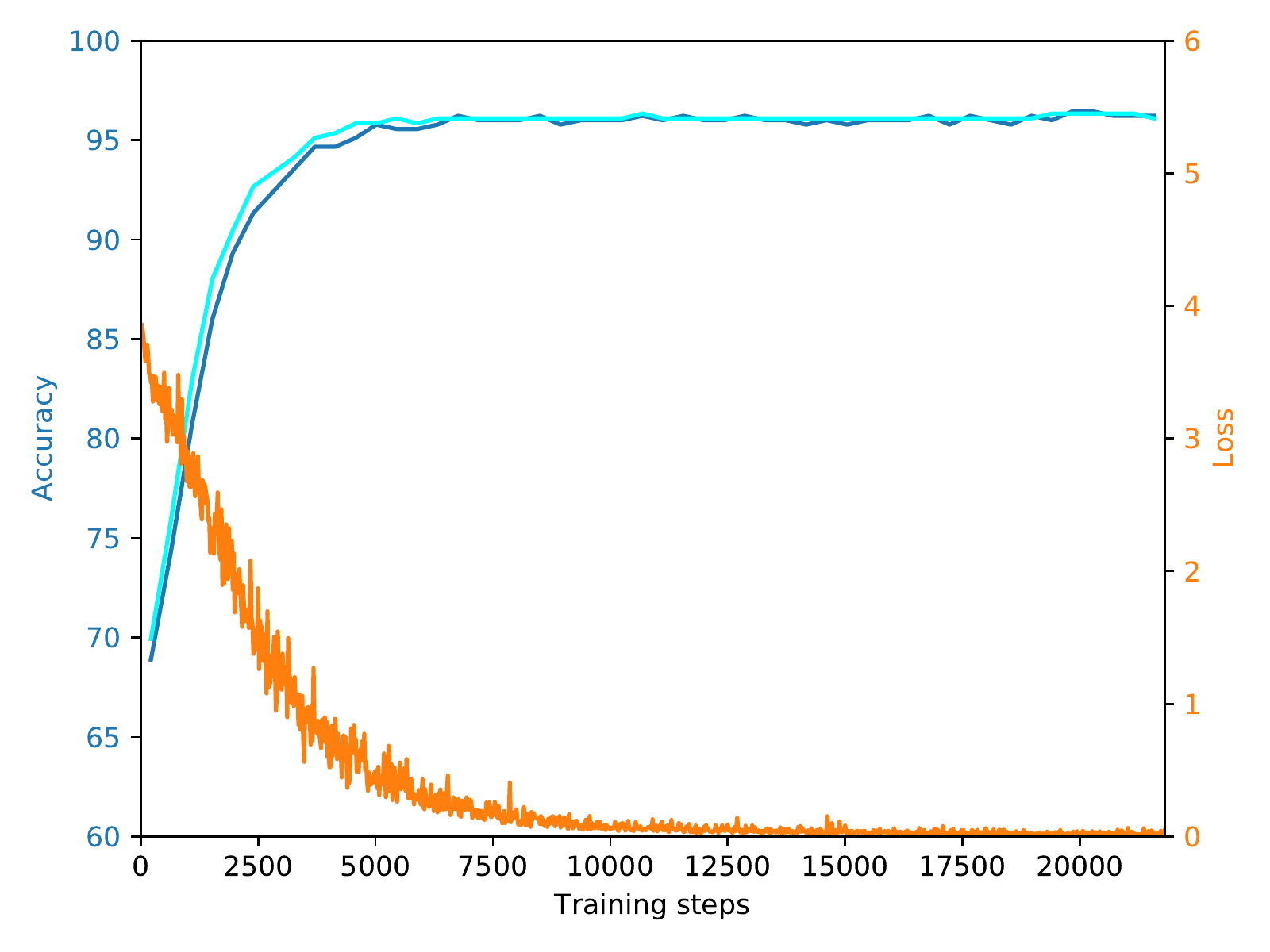}}
    \caption{\textbf{Training-Validation Accuracy and Loss.} Examples of training and validation accuracy over the course of training for 100 epochs for (a) a 50\% open problem and (b) 10\% openness. Cyan: training accuracy, blue: validation accuracy, orange: loss.}
    \label{fig:acc-divergence}
\end{figure}


\subsubsection{Qualitative Analysis}

\begin{figure}[h]
    \centering
    \subfloat[Training\label{subfig:tsne-training}]{ \includegraphics[width=0.32\textwidth]{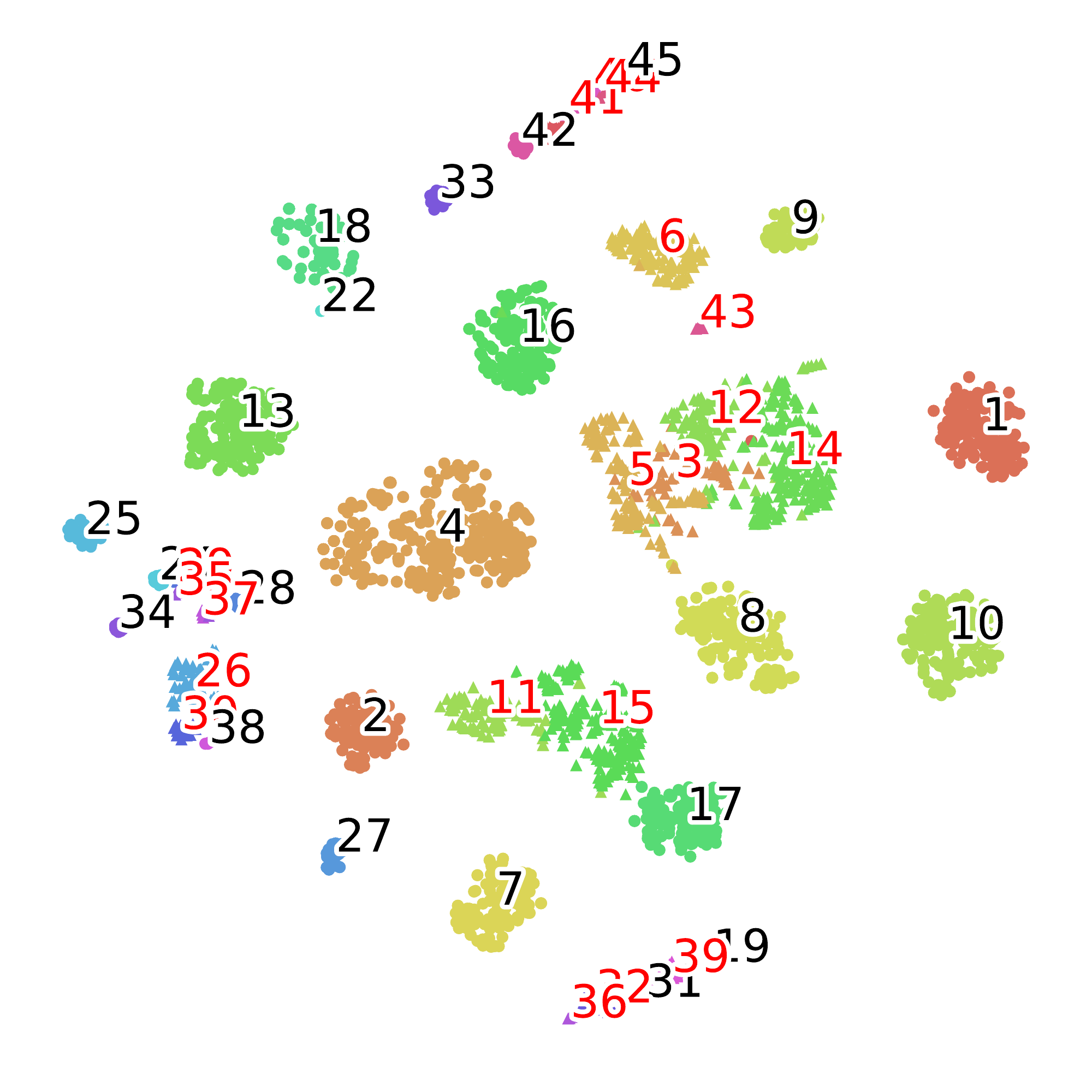}}
    \hfill
    \subfloat[Training\label{subfig:tsne-validation}]{ \includegraphics[width=0.32\textwidth]{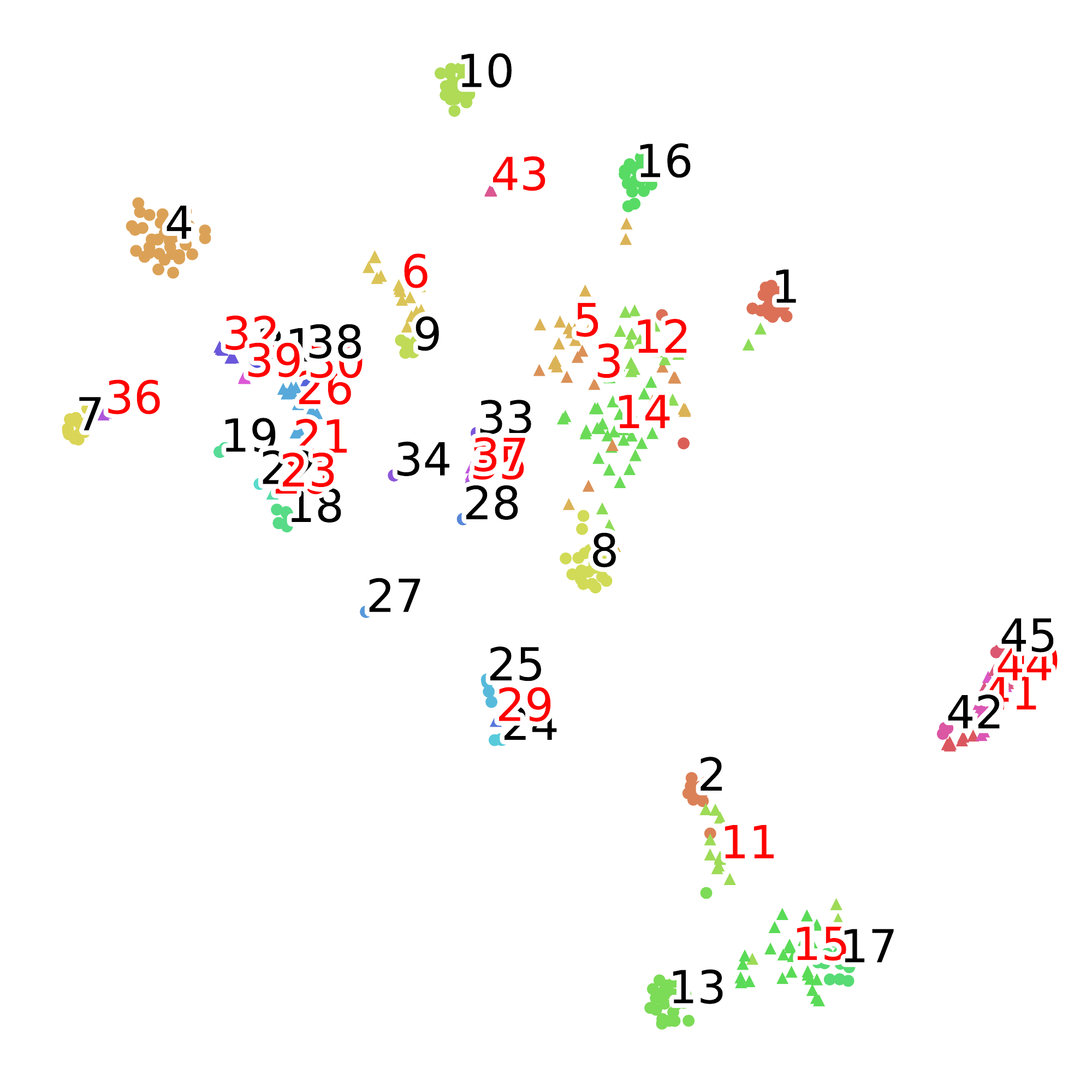}}
    \hfill
    \subfloat[Training\label{subfig:tsne-testing}]{ \includegraphics[width=0.32\textwidth]{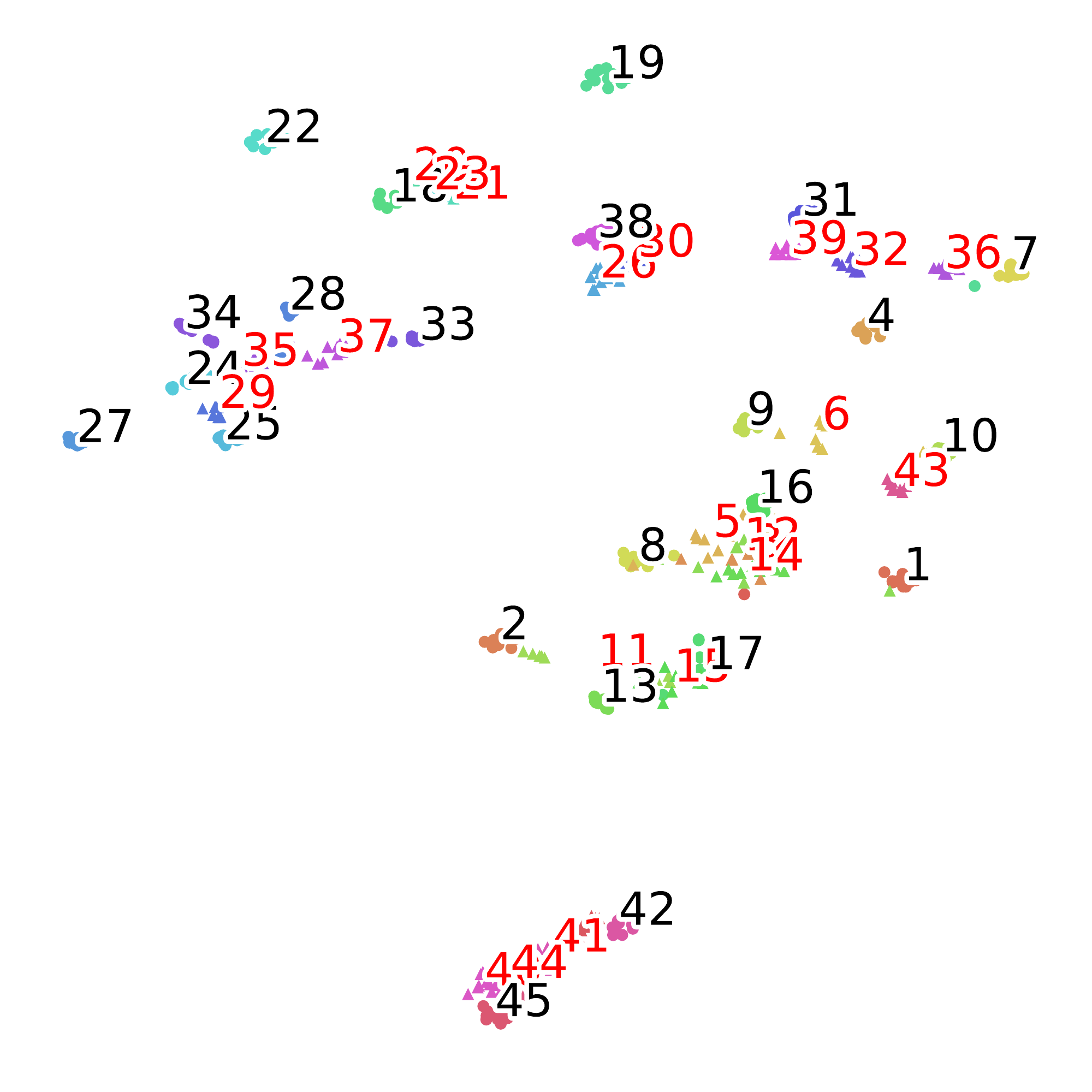}}
    \caption{\textbf{t-SNE \cite{van2008visualizing} Embedding Visualisation}. Examples of the embedded space for training, validation and testing instances for each class. Positions of class labels indicate cluster centroids, with those in red (23 individuals; half of the dataset) denoting unseen classes that were withheld during training. The embedding was trained with our proposed loss function combining a Softmax component with reciprocal triplet loss.}
    \label{fig:tsne-plot}
\end{figure}

To provide a qualitative visualisation, we include Figure~\ref{fig:tsne-plot}, which is a visualisation of the embedded space and the corresponding clusters.
This plot and the others in this section were produced using t-distributed Stochastic Neighbour Embedding (t-SNE) \cite{van2008visualizing}, a technique for visualising high-dimensional spaces, with perplexity of $30$. 
Visible -- particularly in relation to the embedded training set (see Fig.~\ref{subfig:tsne-training}) -- is the success of the model trained via triplet loss formulations, `clumping' like-identities together whilst distancing others.
This is then sufficient to cluster and thereby re-identify never before seen testing identities (see Fig. \ref{subfig:tsne-testing}).
Most importantly in this case, despite only being shown \textit{half} of the identity classes during training, the model learned a discriminative enough embedding that generalises well to previously unseen cattle. 
Thus, surprisingly few coat pattern identities are sufficient to create a latent space that spans dimensions which can successfully accommodate and cluster unseen identities.

\begin{figure}[hbt!]
    \centering
    \subfloat[Triplet Loss]{\includegraphics[width=0.49\textwidth]{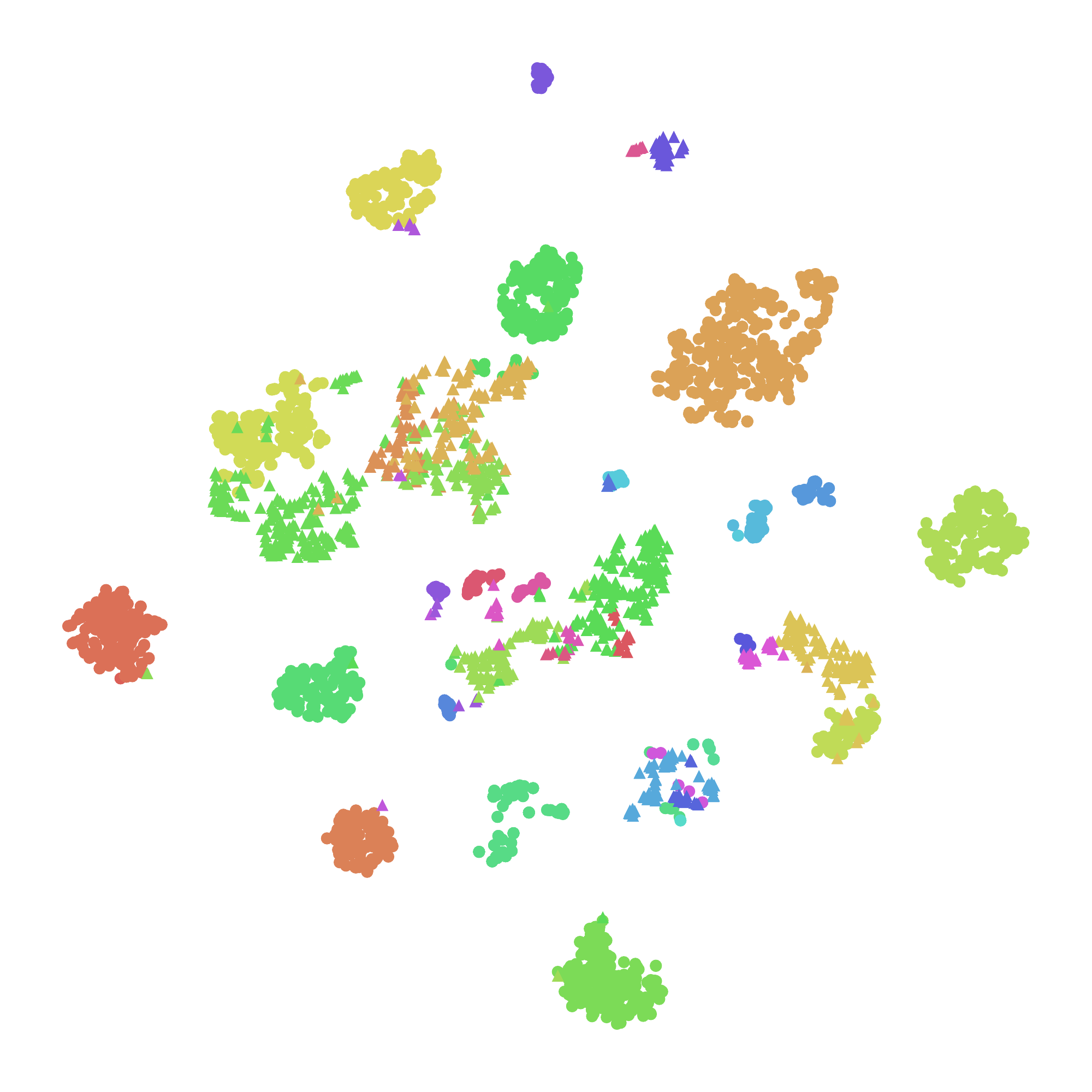}}
    \hfill
    \subfloat[Reciprocal Triplet Loss]{\includegraphics[width=0.49\textwidth]{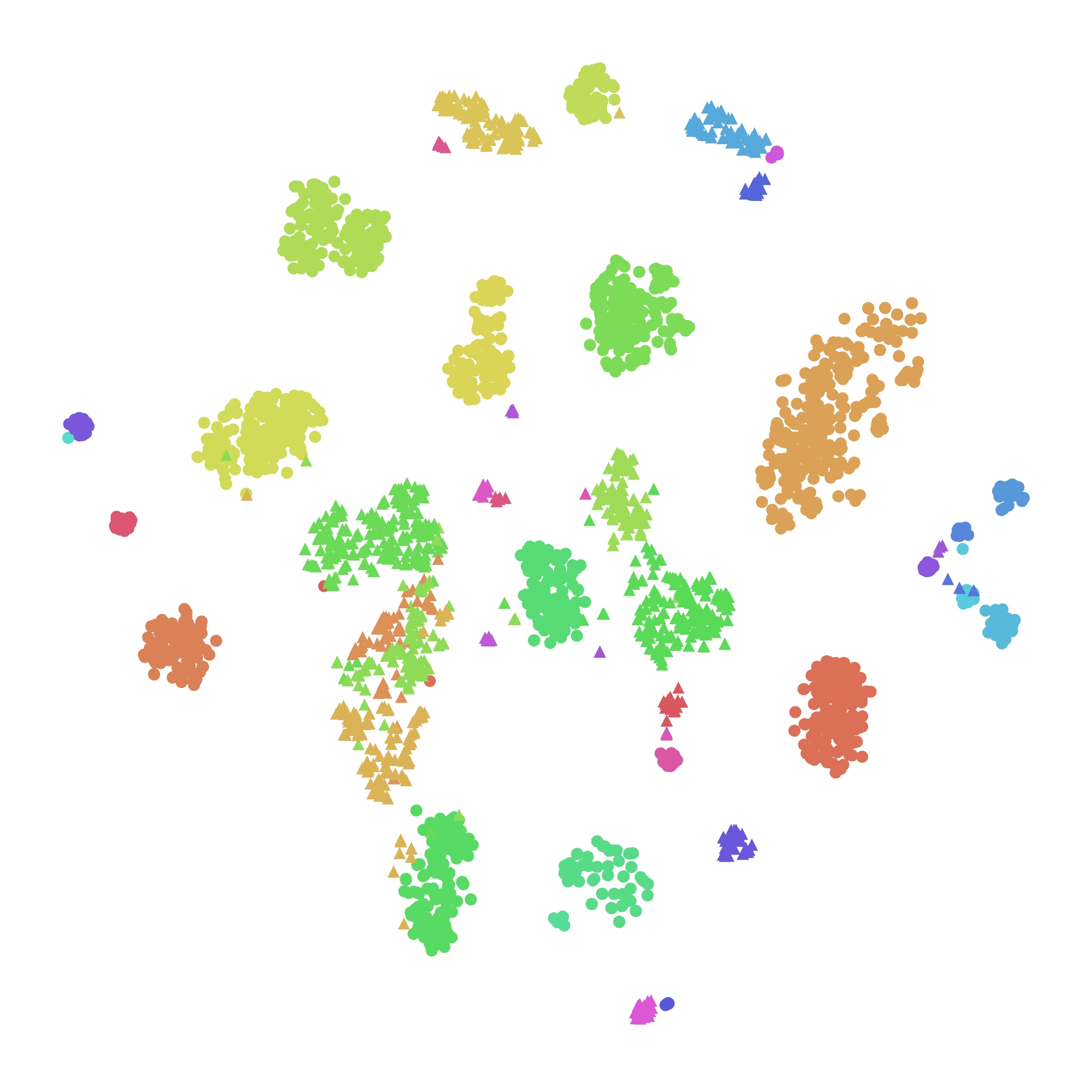}}
    \\
    \subfloat[Softmax + Triplet Loss]{\includegraphics[width=0.49\textwidth]{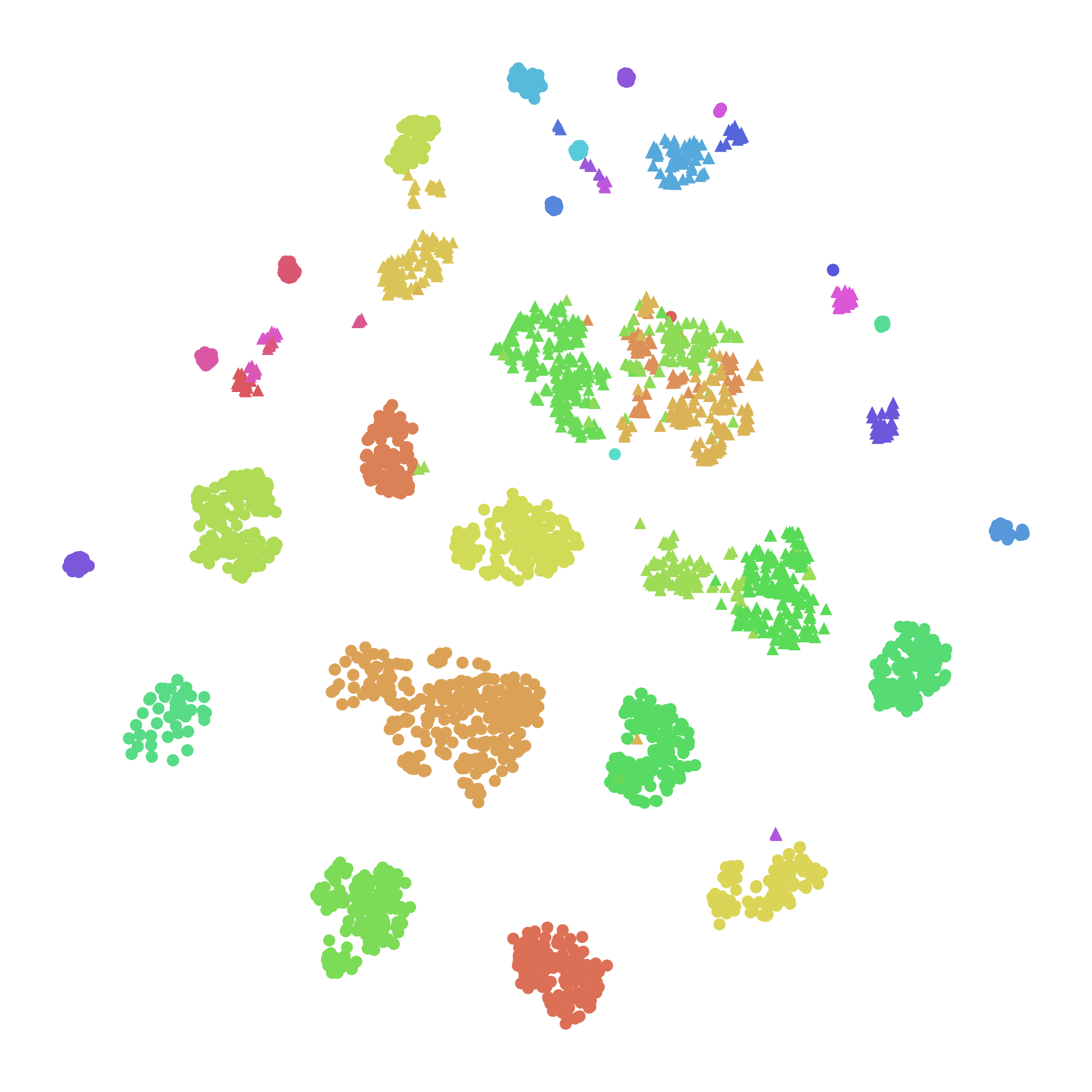}}
    \hfill
    \subfloat[Softmax + Reciprocal Triplet Loss]{\includegraphics[width=0.49\textwidth]{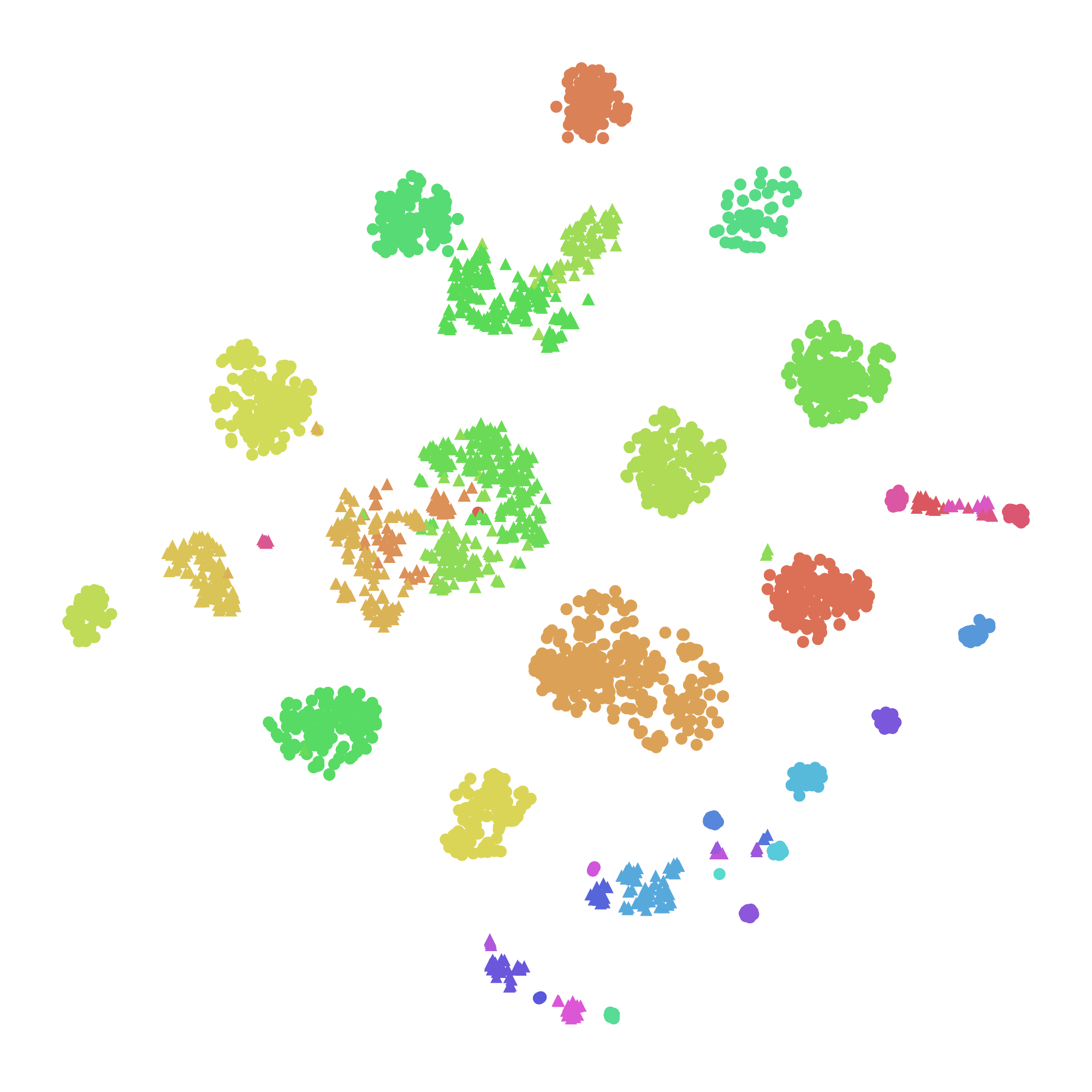}}
    \caption{\textbf{Embeddings Visualisation per Loss Function}. Visualisation of the clusterings for the various loss functions used for training the embedded spaces. The visualisations are for the first fold of the same 50\% open problem across all loss functions. Note that individual colours representing separate classes/animals are consistent across all visualisations. The visualisation is generated using the t-SNE \cite{van2008visualizing} dimensionality reduction technique.}
    \label{fig:embedding-comparison}
\end{figure}

\begin{figure}[hbt!]
    \centering
    \includegraphics[width=0.7\textwidth]{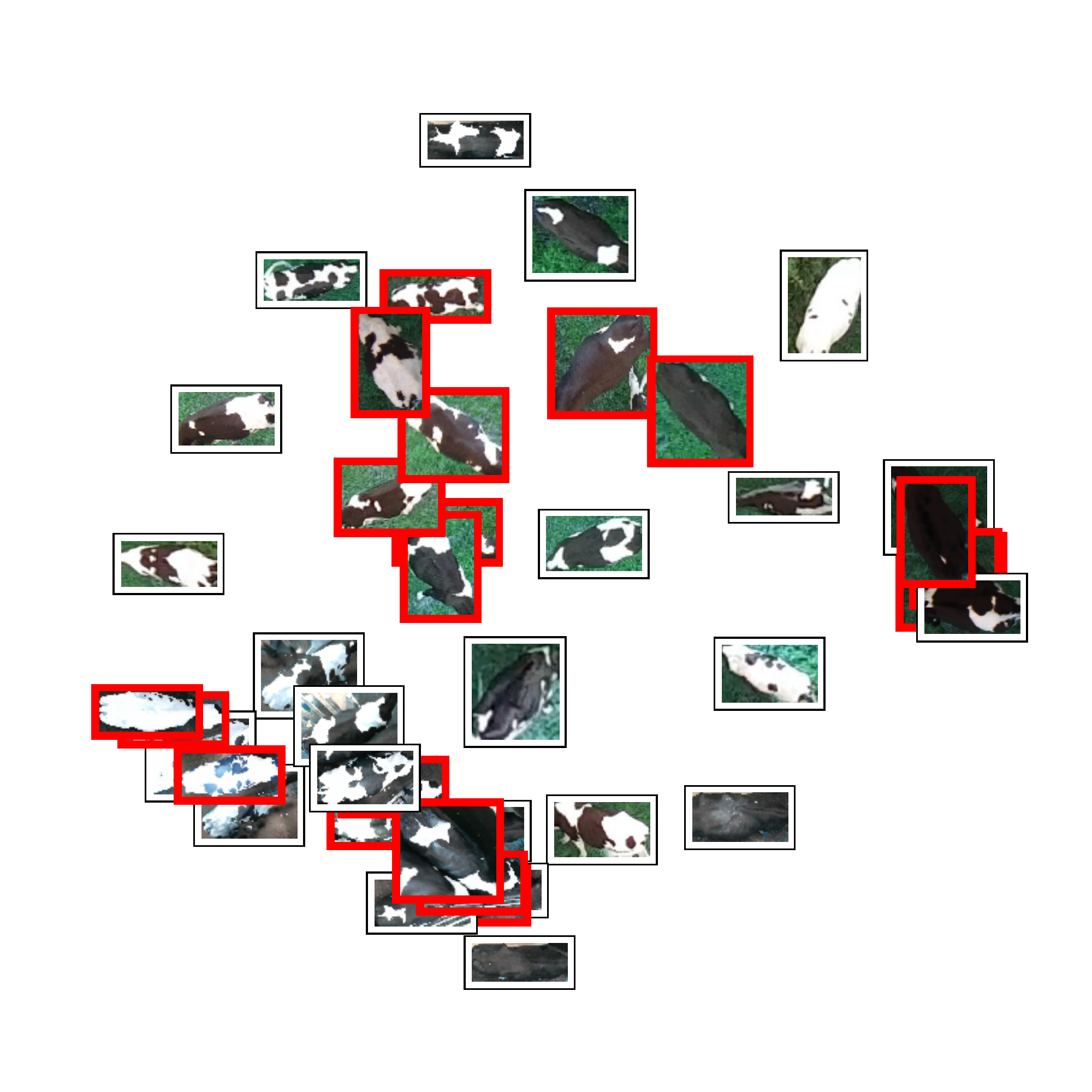}
    \caption{\textbf{Class Examples Overlay.} A randomly chosen example from each class overlaid on the centroids of the embeddings for their respective training instances, where half of the classes (highlighted in red) were not shown during training. Dimensionality reduction from $n=128$ to $2$ was performed using t-SNE \cite{van2008visualizing} and the embedding was trained using Softmax and reciprocal triplet loss.}
    \label{fig:class-overlay}
\end{figure}

Figure \ref{fig:embedding-comparison} visualises the embeddings of the consistent training set for a 50\% open problem across all the implemented loss functions used to the train latent spaces.
The inclusion of a Softmax component in the loss function provided quantifiable improvements in identification accuracy.
This is also reflected in the quality of the embeddings and corresponding clusters, comparing the top and bottom rows in Figure~\ref{fig:embedding-comparison}.
Thus, both quantitative and qualitative findings re-enforce the suitability of the proposed method to the task at hand.
The core technical takeaway is that the inclusion of a fully supervised loss term appears to beneficially support a purely metric learning-based approach in training a discriminative and separable latent representation that is able to generalise to unseen instances of Holstein-Friesians.
Figure \ref{fig:class-overlay} illustrates an example from each class overlaid in this same latent space.
This visualises the spatial similarities and dissimilarities the network uses to generate separable embeddings for the classes that are seen during training that generalise to unseen individuals (shown in red).

To finish, the experiments in this section have relied upon the assumption of good quality cattle regions having been supplied from the previous detection stage - a reasonable assumption given the near perfect performance on the detection task.
In the rare occurrence that this is not the case -- perhaps owing to poor localisation or rare false positive detection -- the proportion of green (for outdoor imagery) and other background pixels increases significantly.
Consequently, we typically found the embeddings of images of this erroneous nature to be distant from other clusters in the latent space.
Thus, the selection of a distance parameter in this space would filter out the vast majority of failures caused by this problem.
This further highlights the suitability of this form of approach in its ability to develop robustness to error cases that are not present in the training set.

\section{Conclusion}
\label{sec:conclusion}

This work proposes a complete pipeline for identifying individual Holstein-Friesian cattle, both seen and never before seen, in agriculturally-relevant imagery.
An assessment of existing state-of-the-art object detectors determined that they are well-suited to serve as an initial breed-wide cattle detector.
Following this, extensive experiments in open-set recognition found that surprisingly few instances are needed in order to learn and construct a robust embedding space -- from image RoI to ID clusters -- that generalises well to unseen cattle.
For instance, for a latent space built from 23 out of 46 individuals using softmax and reciprocal triplet loss, an average accuracy of $93.75\%$ was observed.

Identification of individual cattle is key to dairy farming in most countries, contrasting with other species such as chickens, which are more likely to be treated as a group.
Information gained from monitoring the behaviour and health of individual cows by stockmen is used to make management decisions about that animal, including relating to disease prevention and treatment, fertility and feeding.
Some farms already utilise automated methods of cow management, for example through pedometers to detect activity levels that may indicate oestrus behaviour.
Individual identification of cow location and a range of behaviours from remote observations could provide significant further precision farming opportunities for disease detection and welfare monitoring, including the use of resources aiming to provide positive experiences for cows.
In addition, it could be possible to utilise individual animal monitoring on each farm within a wider context of a network of farms, for example by companies or national agencies, to provide early detection of disease outbreaks and transmission.
Considering its wider application, our work suggests that the proposed pipeline is a viable step towards automating cattle detection and identification non-intrusively in agriculturally-relevant scenarios where herds change dynamically over time.
Importantly, the identification component can be trained at the time of deployment on the current herd and as shown here for the first time, performs well without re-enrolment of individuals or re-training of the system as the population changes - a key requirement for transferability in practical agricultural settings.

\subsection{Future Work}

Further research will look towards tracking from video sequences through continuous re-identification. 
As we have shown that our cattle detection and individual identification techniques are highly accurate, the incorporation of simple tracking techniques between video frames have the potential to filter out any remaining errors. 
How robust this approach will be to heavy bunching of cows (for example, before milking in traditional parlours) remains to be tested.  

Further goals include the incorporation of collision detection for analysis of social networks and transmission dynamics, and behaviour detection for automated welfare and health assessment, which would allow longitudinal tracking of the disease and welfare status of individual cows.
Within this regard, the addition of a depth imagery component alongside standard RGB to support and improve these objectives needs to be evaluated.

We will also look towards investigating the scalability of our approach to large populations. 
That is, increasing the base number of individuals via additional data acquisition with the intention of learning a general representation of dorsal features exhibited by Holstein-Friesian cattle. 
In doing so, this paves the way for the model to generalise to new farms and new herds prior to deployment without any training, with significant implications for the precision livestock farming sector.


\section*{Acknowledgements}
This work was supported by The Alan Turing Institute under the EPSRC grant EP/N510129/1 and the John Oldacre Foundation through the John Oldacre Centre for Sustainability and Welfare in Dairy Production, Bristol Veterinary School. We thank Suzanne Held, David Barrett, and Mike Mendl of Bristol Veterinary School for fruitful discussions and suggestions, and Kate Robinson and the Wyndhurst Farm staff for their assistance with data collection. Thanks also to Miguel Lagunes-Fortiz for permitting use, adaptation and redistribution of key source code.

\bibliography{mybibfile}

\begin{thebibliography}{10}
\expandafter\ifx\csname url\endcsname\relax
  \def\url#1{\texttt{#1}}\fi
\expandafter\ifx\csname urlprefix\endcsname\relax\def\urlprefix{URL }\fi
\expandafter\ifx\csname href\endcsname\relax
  \def\href#1#2{#2} \def\path#1{#1}\fi

\bibitem{sourcecodedatasets}
Source code and network weights:
  \url{https://github.com/CWOA/MetricLearningIdentification}, OpenCows2020
  dataset: \url{https://doi.org/10.5523/bris.10m32xl88x2b61zlkkgz3fml17}.

\bibitem{whff}
W.~H.~F. Federation, Annual statistics,
  \url{http://www.whff.info/documentation/statistics.php}, [Online; accessed
  4-August-2020].

\bibitem{tadesse2003milk}
M.~Tadesse, T.~Dessie, Milk production performance of zebu, holstein friesian
  and their crosses in ethiopia, Livestock Research for Rural Development
  15~(3) (2003) 1--9.

\bibitem{faocattle}
Food, A.~O. of~the United~Nations, Gateway to dairy production and products,
  \url{http://www.fao.org/dairy-production-products/production/dairy-animals/cattle/en/},
  [Online; accessed 4-August-2020].

\bibitem{faostat}
Food, A.~O. of~the United~Nations, Faostat,
  \url{http://www.fao.org/faostat/en/#data/QL}, [Online; accessed
  4-August-2020].

\bibitem{velez2013beef}
J.~Velez, A.~Sanchez, J.~Sanchez, J.~Esteban, Beef identification in industrial
  slaughterhouses using machine vision techniques, Spanish Journal of
  Agricultural Research 11~(4) (2013) 945--957.

\bibitem{pennington2007tattooing}
J.~A. Pennington, Tattooing of cattle and goats, Cooperative Extension Service,
  University of Arkansas Division of~…, 2007.

\bibitem{bertram1996freeze}
L.~Bertram, B.~Gill, J.~Coventry, A.~Springs, F.~DPIFM, Freeze branding,
  Department of industrires and fisheries, northern territory, Darwin,
  Australia.

\bibitem{eu82097}
E.~Parliament, Council, Establishing a system for the identification and
  registration of bovine animals and regarding the labelling of beef and beef
  products and repealing council regulation (ec) no 820/97,
  \url{http://eur-lex.europa.eu/legal-content/EN/TXT/?uri=celex:32000R1760},
  [Online; accessed 29-January-2016] (1997).

\bibitem{united2018states}
U.~S.~D. of~Agriculture (USDA)~Animal, P.~H.~I. Service, Cattle identification,
  \url{https://www.aphis.usda.gov/aphis/ourfocus/animalhealth/nvap/NVAP-Reference-Guide/Animal-Identification/Cattle-Identification},
  [Online; accessed 14-November-2018].

\bibitem{hansen2018automated}
M.~F. Hansen, M.~L. Smith, L.~N. Smith, K.~A. Jabbar, D.~Forbes, Automated
  monitoring of dairy cow body condition, mobility and weight using a single 3d
  video capture device, Computers in industry 98 (2018) 14--22.

\bibitem{smith2005traceability}
G.~Smith, J.~Tatum, K.~Belk, J.~Scanga, T.~Grandin, J.~Sofos, Traceability from
  a us perspective, Meat science 71~(1) (2005) 174--193.

\bibitem{bowling2008identification}
M.~Bowling, D.~Pendell, D.~Morris, Y.~Yoon, K.~Katoh, K.~Belk, G.~Smith,
  Identification and traceability of cattle in selected countries outside of
  north america, The Professional Animal Scientist 24~(4) (2008) 287--294.

\bibitem{caporale2001importance}
V.~Caporale, A.~Giovannini, C.~Di~Francesco, P.~Calistri, Importance of the
  traceability of animals and animal products in epidemiology, Revue
  Scientifique et Technique-Office International des Epizooties 20~(2) (2001)
  372--378.

\bibitem{houston2001computerised}
R.~Houston, A computerised database system for bovine traceability, Revue
  Scientifique et Technique-Office International des Epizooties 20~(2) (2001)
  652.

\bibitem{buick2004animal}
W.~Buick, Animal passports and identification, Defra Veterinary Journal 15
  (2004) 20--26.

\bibitem{shanahan2009framework}
C.~Shanahan, B.~Kernan, G.~Ayalew, K.~McDonnell, F.~Butler, S.~Ward, A
  framework for beef traceability from farm to slaughter using global
  standards: an irish perspective, Computers and electronics in agriculture
  66~(1) (2009) 62--69.

\bibitem{klindtworth1999electronic}
M.~Klindtworth, G.~Wendl, K.~Klindtworth, H.~Pirkelmann, Electronic
  identification of cattle with injectable transponders, Computers and
  electronics in agriculture 24~(1-2) (1999) 65--79.

\bibitem{adcock2018branding}
S.~J. Adcock, C.~B. Tucker, G.~Weerasinghe, E.~Rajapaksha, Branding practices
  on four dairies in kantale, sri lanka, Animals 8~(8) (2018) 137.

\bibitem{awad2016classical}
A.~I. Awad, From classical methods to animal biometrics: A review on cattle
  identification and tracking, Computers and Electronics in Agriculture 123
  (2016) 423--435.

\bibitem{ungar2005inference}
E.~D. Ungar, Z.~Henkin, M.~Gutman, A.~Dolev, A.~Genizi, D.~Ganskopp, Inference
  of animal activity from gps collar data on free-ranging cattle, Rangeland
  Ecology \& Management 58~(3) (2005) 256--266.

\bibitem{turner2000monitoring}
L.~Turner, M.~Udal, B.~Larson, S.~Shearer, Monitoring cattle behavior and
  pasture use with gps and gis, Canadian Journal of Animal Science 80~(3)
  (2000) 405--413.

\bibitem{johnston19961418001}
A.~Johnston, D.~Edwards, E.~Hofmann, P.~Wrench, F.~Sharples, R.~Hiller,
  W.~Welte, K.~Diederichs, 1418001. welfare implications of identification of
  cattle by ear tags, The Veterinary Record 138~(25) (1996) 612--614.

\bibitem{edwards1999welfare}
D.~Edwards, A.~Johnston, Welfare implications of sheep ear tags., The
  Veterinary Record 144~(22) (1999) 603--606.

\bibitem{fosgate2006ear}
G.~Fosgate, A.~Adesiyun, D.~Hird, Ear-tag retention and identification methods
  for extensively managed water buffalo (bubalus bubalis) in trinidad,
  Preventive veterinary medicine 73~(4) (2006) 287--296.

\bibitem{edwards2001comparison}
D.~Edwards, A.~Johnston, D.~Pfeiffer, A comparison of commonly used ear tags on
  the ear damage of sheep, Animal Welfare 10~(2) (2001) 141--151.

\bibitem{wardrope1995problems}
D.~Wardrope, Problems with the use of ear tags in cattle, Veterinary Record
  137~(26) (1995) 675--675.

\bibitem{martinez2013video}
C.~A. Martinez-Ortiz, R.~M. Everson, T.~Mottram, Video tracking of dairy cows
  for assessing mobility scores.

\bibitem{li2017automatic}
W.~Li, Z.~Ji, L.~Wang, C.~Sun, X.~Yang, Automatic individual identification of
  holstein dairy cows using tailhead images, Computers and electronics in
  agriculture 142 (2017) 622--631.

\bibitem{andrew2016automatic}
W.~Andrew, S.~Hannuna, N.~Campbell, T.~Burghardt, Automatic individual holstein
  friesian cattle identification via selective local coat pattern matching in
  rgb-d imagery, in: 2016 IEEE International Conference on Image Processing
  (ICIP), IEEE, 2016, pp. 484--488.

\bibitem{andrew2017visual}
W.~Andrew, C.~Greatwood, T.~Burghardt, Visual localisation and individual
  identification of holstein friesian cattle via deep learning, in: Proceedings
  of the IEEE International Conference on Computer Vision, 2017, pp.
  2850--2859.

\bibitem{andrew2019aerial}
W.~Andrew, C.~Greatwood, T.~Burghardt, Aerial animal biometrics: Individual
  friesian cattle recovery and visual identification via an autonomous uav with
  onboard deep inference, in: 2019 IEEE/RSJ International Conference on
  Intelligent Robots and Systems (IROS), IEEE, 2019, pp. 237--243.

\bibitem{andrew2019visual}
W.~Andrew, Visual biometric processes for collective identification of
  individual friesian cattle, Ph.D. thesis, University of Bristol (2019).

\bibitem{kuhl2013}
H.~S. K{\"u}hl, T.~Burghardt, Animal biometrics: quantifying and detecting
  phenotypic appearance, Trends in ecology \& evolution 28~(7) (2013) 432--441.

\bibitem{petersen1922identification}
W.~Petersen, The identification of the bovine by means of nose-prints, Journal
  of dairy science 5~(3) (1922) 249--258.

\bibitem{kumar2017automatic}
S.~Kumar, S.~K. Singh, Automatic identification of cattle using muzzle point
  pattern: a hybrid feature extraction and classification paradigm, Multimedia
  Tools and Applications 76~(24) (2017) 26551--26580.

\bibitem{kumar2017real}
S.~Kumar, S.~K. Singh, R.~S. Singh, A.~K. Singh, S.~Tiwari, Real-time
  recognition of cattle using animal biometrics, Journal of Real-Time Image
  Processing 13~(3) (2017) 505--526.

\bibitem{kimura2004structural}
A.~Kimura, K.~Itaya, T.~Watanabe, Structural pattern recognition of biological
  textures with growing deformations: A case of cattle's muzzle patterns,
  Electronics and Communications in Japan (Part II: Electronics) 87~(5) (2004)
  54--66.

\bibitem{tharwat2014cattle}
A.~Tharwat, T.~Gaber, A.~E. Hassanien, H.~A. Hassanien, M.~F. Tolba, Cattle
  identification using muzzle print images based on texture features approach,
  in: Proceedings of the Fifth International Conference on Innovations in
  Bio-Inspired Computing and Applications IBICA 2014, Springer, 2014, pp.
  217--227.

\bibitem{awad2019bag}
A.~I. Awad, M.~Hassaballah, Bag-of-visual-words for cattle identification from
  muzzle print images, Applied Sciences 9~(22) (2019) 4914.

\bibitem{el2015bovines}
H.~M. El~Hadad, H.~A. Mahmoud, F.~A. Mousa, Bovines muzzle classification based
  on machine learning techniques, Procedia Computer Science 65 (2015) 864--871.

\bibitem{barry2007using}
B.~Barry, U.~Gonzales-Barron, K.~McDonnell, F.~Butler, S.~Ward, Using muzzle
  pattern recognition as a biometric approach for cattle identification,
  Transactions of the ASABE 50~(3) (2007) 1073--1080.

\bibitem{allen2008evaluation}
A.~Allen, B.~Golden, M.~Taylor, D.~Patterson, D.~Henriksen, R.~Skuce,
  Evaluation of retinal imaging technology for the biometric identification of
  bovine animals in northern ireland, Livestock science 116~(1-3) (2008)
  42--52.

\bibitem{barbedo2019study}
J.~G.~A. Barbedo, L.~V. Koenigkan, T.~T. Santos, P.~M. Santos, A study on the
  detection of cattle in uav images using deep learning, Sensors 19~(24) (2019)
  5436.

\bibitem{cai2013cattle}
C.~Cai, J.~Li, Cattle face recognition using local binary pattern descriptor,
  in: 2013 Asia-Pacific Signal and Information Processing Association Annual
  Summit and Conference, IEEE, 2013, pp. 1--4.

\bibitem{arslan20143d}
A.~C. Arslan, M.~Akar, F.~Alag{\"o}z, 3d cow identification in cattle farms,
  in: 2014 22nd Signal Processing and Communications Applications Conference
  (SIU), IEEE, 2014, pp. 1347--1350.

\bibitem{hu2020cow}
H.~Hu, B.~Dai, W.~Shen, X.~Wei, J.~Sun, R.~Li, Y.~Zhang, Cow identification
  based on fusion of deep parts features, Biosystems Engineering 192 (2020)
  245--256.

\bibitem{qiao2019individual}
Y.~Qiao, D.~Su, H.~Kong, S.~Sukkarieh, S.~Lomax, C.~Clark, Individual cattle
  identification using a deep learning based framework, IFAC-PapersOnLine
  52~(30) (2019) 318--323.

\bibitem{bhole2019computer}
A.~Bhole, O.~Falzon, M.~Biehl, G.~Azzopardi, A computer vision pipeline that
  uses thermal and rgb images for the recognition of holstein cattle, in:
  International Conference on Computer Analysis of Images and Patterns,
  Springer, 2019, pp. 108--119.

\bibitem{liu2016ssd}
W.~Liu, D.~Anguelov, D.~Erhan, C.~Szegedy, S.~Reed, C.-Y. Fu, A.~C. Berg, Ssd:
  Single shot multibox detector, in: European conference on computer vision,
  Springer, 2016, pp. 21--37.

\bibitem{Redmon_2016_CVPR}
J.~Redmon, S.~Divvala, R.~Girshick, A.~Farhadi, You only look once: Unified,
  real-time object detection, in: The IEEE Conference on Computer Vision and
  Pattern Recognition (CVPR), 2016.

\bibitem{ren2015faster}
S.~Ren, K.~He, R.~Girshick, J.~Sun, Faster r-cnn: Towards real-time object
  detection with region proposal networks, in: Advances in neural information
  processing systems, 2015, pp. 91--99.

\bibitem{cai2018cascade}
Z.~Cai, N.~Vasconcelos, Cascade r-cnn: Delving into high quality object
  detection, in: Proceedings of the IEEE conference on computer vision and
  pattern recognition, 2018, pp. 6154--6162.

\bibitem{redmon2018yolov3}
J.~Redmon, A.~Farhadi, Yolov3: An incremental improvement, arXiv preprint
  arXiv:1804.02767.

\bibitem{lin2017focal}
T.-Y. Lin, P.~Goyal, R.~Girshick, K.~He, P.~Doll{\'a}r, Focal loss for dense
  object detection, in: Proceedings of the IEEE international conference on
  computer vision, 2017, pp. 2980--2988.

\bibitem{jain2014multi}
L.~P. Jain, W.~J. Scheirer, T.~E. Boult, Multi-class open set recognition using
  probability of inclusion, in: European Conference on Computer Vision,
  Springer, 2014, pp. 393--409.

\bibitem{scheirer2014probability}
W.~J. Scheirer, L.~P. Jain, T.~E. Boult, Probability models for open set
  recognition, IEEE transactions on pattern analysis and machine intelligence
  36~(11) (2014) 2317--2324.

\bibitem{rudd2017extreme}
E.~M. Rudd, L.~P. Jain, W.~J. Scheirer, T.~E. Boult, The extreme value machine,
  IEEE transactions on pattern analysis and machine intelligence 40~(3) (2017)
  762--768.

\bibitem{scheirer2012toward}
W.~J. Scheirer, A.~de~Rezende~Rocha, A.~Sapkota, T.~E. Boult, Toward open set
  recognition, IEEE transactions on pattern analysis and machine intelligence
  35~(7) (2012) 1757--1772.

\bibitem{junior2016specialized}
P.~R.~M. J{\'u}nior, T.~E. Boult, J.~Wainer, A.~Rocha, Specialized support
  vector machines for open-set recognition, arXiv preprint arXiv:1606.03802.

\bibitem{bendale2015towards}
A.~Bendale, T.~Boult, Towards open world recognition, in: Proceedings of the
  IEEE Conference on Computer Vision and Pattern Recognition, 2015, pp.
  1893--1902.

\bibitem{junior2017nearest}
P.~R.~M. J{\'u}nior, R.~M. de~Souza, R.~d.~O. Werneck, B.~V. Stein, D.~V.
  Pazinato, W.~R. de~Almeida, O.~A. Penatti, R.~d.~S. Torres, A.~Rocha, Nearest
  neighbors distance ratio open-set classifier, Machine Learning 106~(3) (2017)
  359--386.

\bibitem{sermanet2013overfeat}
P.~Sermanet, D.~Eigen, X.~Zhang, M.~Mathieu, R.~Fergus, Y.~LeCun, Overfeat:
  Integrated recognition, localization and detection using convolutional
  networks, arXiv preprint arXiv:1312.6229.

\bibitem{girshick2014rich}
R.~Girshick, J.~Donahue, T.~Darrell, J.~Malik, Rich feature hierarchies for
  accurate object detection and semantic segmentation, in: Proceedings of the
  IEEE conference on computer vision and pattern recognition, 2014, pp.
  580--587.

\bibitem{krizhevsky2012imagenet}
A.~Krizhevsky, I.~Sutskever, G.~E. Hinton, Imagenet classification with deep
  convolutional neural networks, in: Advances in neural information processing
  systems, 2012, pp. 1097--1105.

\bibitem{oza2019deep}
P.~Oza, V.~M. Patel, Deep cnn-based multi-task learning for open-set
  recognition, arXiv preprint arXiv:1903.03161.

\bibitem{yoshihashi2019classification}
R.~Yoshihashi, W.~Shao, R.~Kawakami, S.~You, M.~Iida, T.~Naemura,
  Classification-reconstruction learning for open-set recognition, in:
  Proceedings of the IEEE Conference on Computer Vision and Pattern
  Recognition, 2019, pp. 4016--4025.

\bibitem{bendale2016towards}
A.~Bendale, T.~E. Boult, Towards open set deep networks, in: Proceedings of the
  IEEE conference on computer vision and pattern recognition, 2016, pp.
  1563--1572.

\bibitem{neal2018open}
L.~Neal, M.~Olson, X.~Fern, W.-K. Wong, F.~Li, Open set learning with
  counterfactual images, in: Proceedings of the European Conference on Computer
  Vision (ECCV), 2018, pp. 613--628.

\bibitem{ge2017generative}
Z.~Ge, S.~Demyanov, Z.~Chen, R.~Garnavi, Generative openmax for multi-class
  open set classification, arXiv preprint arXiv:1707.07418.

\bibitem{shu2017doc}
L.~Shu, H.~Xu, B.~Liu, Doc: Deep open classification of text documents, arXiv
  preprint arXiv:1709.08716.

\bibitem{geng2018recent}
C.~Geng, S.-j. Huang, S.~Chen, Recent advances in open set recognition: A
  survey, arXiv preprint arXiv:1811.08581.

\bibitem{meyer2019importance}
B.~J. Meyer, T.~Drummond, The importance of metric learning for robotic vision:
  Open set recognition and active learning, in: 2019 International Conference
  on Robotics and Automation (ICRA), IEEE, 2019, pp. 2924--2931.

\bibitem{lagunes2019learning}
M.~Lagunes-Fortiz, D.~Damen, W.~Mayol-Cuevas, Learning discriminative
  embeddings for object recognition on-the-fly, in: 2019 International
  Conference on Robotics and Automation (ICRA), IEEE, 2019, pp. 2932--2938.

\bibitem{hassen2018learning}
M.~Hassen, P.~K. Chan, Learning a neural-network-based representation for open
  set recognition, arXiv preprint arXiv:1802.04365.

\bibitem{schroff2015facenet}
F.~Schroff, D.~Kalenichenko, J.~Philbin, Facenet: A unified embedding for face
  recognition and clustering, in: Proceedings of the IEEE conference on
  computer vision and pattern recognition, 2015, pp. 815--823.

\bibitem{hermans2017defense}
A.~Hermans, L.~Beyer, B.~Leibe, In defense of the triplet loss for person
  re-identification, arXiv preprint arXiv:1703.07737.

\bibitem{oh2016deep}
H.~Oh~Song, Y.~Xiang, S.~Jegelka, S.~Savarese, Deep metric learning via lifted
  structured feature embedding, in: Proceedings of the IEEE conference on
  computer vision and pattern recognition, 2016, pp. 4004--4012.

\bibitem{opitz2018deep}
M.~Opitz, G.~Waltner, H.~Possegger, H.~Bischof, Deep metric learning with bier:
  Boosting independent embeddings robustly, IEEE transactions on pattern
  analysis and machine intelligence.

\bibitem{oh2017deep}
H.~Oh~Song, S.~Jegelka, V.~Rathod, K.~Murphy, Deep metric learning via facility
  location, in: Proceedings of the IEEE Conference on Computer Vision and
  Pattern Recognition, 2017, pp. 5382--5390.

\bibitem{masullo2019goes}
A.~Masullo, T.~Burghardt, D.~Damen, T.~Perrett, M.~Mirmehdi, Who goes there?
  exploiting silhouettes and wearable signals for subject identification in
  multi-person environments, in: Proceedings of the IEEE International
  Conference on Computer Vision Workshops, 2019, pp. 0--0.

\bibitem{pascal-voc-2012}
M.~Everingham, L.~Van~Gool, C.~K.~I. Williams, J.~Winn, A.~Zisserman, The
  {PASCAL} {V}isual {O}bject {C}lasses {C}hallenge 2012 {(VOC2012)} {R}esults,
  http://www.pascal-network.org/challenges/VOC/voc2012/workshop/index.html.

\bibitem{lin2017feature}
T.-Y. Lin, P.~Doll{\'a}r, R.~Girshick, K.~He, B.~Hariharan, S.~Belongie,
  Feature pyramid networks for object detection, in: Proceedings of the IEEE
  conference on computer vision and pattern recognition, 2017, pp. 2117--2125.

\bibitem{he2016deep}
K.~He, X.~Zhang, S.~Ren, J.~Sun, Deep residual learning for image recognition,
  in: Proceedings of the IEEE conference on computer vision and pattern
  recognition, 2016, pp. 770--778.

\bibitem{deng2009imagenet}
J.~Deng, W.~Dong, R.~Socher, L.-J. Li, K.~Li, L.~Fei-Fei, Imagenet: A
  large-scale hierarchical image database, in: 2009 IEEE conference on computer
  vision and pattern recognition, Ieee, 2009, pp. 248--255.

\bibitem{robbins1951stochastic}
H.~Robbins, S.~Monro, A stochastic approximation method, The annals of
  mathematical statistics (1951) 400--407.

\bibitem{qian1999momentum}
N.~Qian, On the momentum term in gradient descent learning algorithms, Neural
  networks 12~(1) (1999) 145--151.

\bibitem{stephen1990perceptron}
I.~Stephen, Perceptron-based learning algorithms, IEEE Transactions on neural
  networks 50~(2) (1990) 179.

\bibitem{lin2014microsoft}
T.-Y. Lin, M.~Maire, S.~Belongie, J.~Hays, P.~Perona, D.~Ramanan,
  P.~Doll{\'a}r, C.~L. Zitnick, Microsoft coco: Common objects in context, in:
  European conference on computer vision, Springer, 2014, pp. 740--755.

\bibitem{hadsell2006dimensionality}
R.~Hadsell, S.~Chopra, Y.~LeCun, Dimensionality reduction by learning an
  invariant mapping, in: 2006 IEEE Computer Society Conference on Computer
  Vision and Pattern Recognition (CVPR'06), Vol.~2, IEEE, 2006, pp. 1735--1742.

\bibitem{hodan2017t}
T.~Hodan, P.~Haluza, {\v{S}}.~Obdr{\v{z}}{\'a}lek, J.~Matas, M.~Lourakis,
  X.~Zabulis, T-less: An rgb-d dataset for 6d pose estimation of texture-less
  objects, in: 2017 IEEE Winter Conference on Applications of Computer Vision
  (WACV), IEEE, 2017, pp. 880--888.

\bibitem{wang2017object}
X.~Wang, F.~M. Eliott, J.~Ainooson, J.~H. Palmer, M.~Kunda, An object is worth
  six thousand pictures: The egocentric, manual, multi-image (emmi) dataset,
  in: Proceedings of the IEEE International Conference on Computer Vision,
  2017, pp. 2364--2372.

\bibitem{balntas2016learning}
V.~Balntas, E.~Riba, D.~Ponsa, K.~Mikolajczyk, Learning local feature
  descriptors with triplets and shallow convolutional neural networks., in:
  Bmvc, Vol.~1, 2016, p.~3.

\bibitem{van2008visualizing}
L.~J. van~der Maaten, G.~E. Hinton, Visualizing high-dimensional data using
  t-sne, Journal of machine learning research 9~(nov) (2008) 2579--2605.

\end{thebibliography}

\end{document}